\documentclass[journal]{IEEEtran}
\ifCLASSINFOpdf
\else
\fi
\usepackage{amsmath}
\usepackage{array}

\usepackage{algorithm}
\usepackage{algpseudocode}
\usepackage{graphicx}
\usepackage{subfigure}
\usepackage{booktabs}
\usepackage{multirow}
\usepackage{txfonts}

\begin{document}
%
\title{Hyperspectral Mixed Noise Removal via Subspace Representation and Weighted Low-rank Tensor Regularization}
%
%
%

\author{Hang Zhou,~\IEEEmembership{}
        Yanchi Su,~\IEEEmembership{}
        Zhanshan Li~\IEEEmembership{}
\thanks{Hang Zhou is with the Department
of Software Engineering, Jilin University, Changchun (e-mail: zhouhang19970714@163.com).}
\thanks{Yanchi Su is with the Department
of Artificial Intelligence, Jilin University, Changchun (e-mail: suyanchi@gmail.com).}
\thanks{Zhanshan Li is with the Department
of Software Engineering, Jilin University, Changchun (e-mail: zslizsli@163.com). Corresponding author.}
}

%
%

\markboth{Journal of \LaTeX\ Class Files}%
{Zhou \MakeLowercase{\textit{et al.}}: Hyperspectral Mixed Noise Removal via Subspace Representation and Weighted Low-rank Tensor Regularization}
%



\maketitle

\begin{abstract}
Recently, the low-rank property of different components extracted from the image has been considered in many hyperspectral image denoising methods. However, these methods usually unfold the 3D tensor to 2D matrix or 1D vector to exploit the prior information, such as nonlocal spatial self-similarity (NSS) and global spectral correlation (GSC), which break the intrinsic structure correlation of hyperspectral image (HSI) and thus lead to poor restoration quality. In addition, most of them suffer from heavy computational burden issues due to the involvement of singular value decomposition operation on matrix and tensor in the original high-dimensionality space of HSI. We employ subspace representation and the weighted low-rank tensor regularization (SWLRTR) into the model to remove the mixed noise in the hyperspectral image. Specifically, to employ the GSC among spectral bands, the noisy HSI is projected into a low-dimensional subspace which simplified calculation. After that, a weighted low-rank tensor regularization term is introduced to characterize the priors in the reduced image subspace. Moreover, we design an algorithm based on alternating minimization to solve the nonconvex problem. Experiments on simulated and real datasets demonstrate that the SWLRTR method performs better than other hyperspectral denoising methods quantitatively and visually.
\end{abstract}


%
\IEEEpeerreviewmaketitle

\section{Introduction}
%
%
%
%
\IEEEPARstart{T}{hanks}
to the advancements in imaging technology, hyperspectral image (HSI) is capable of providing abundant information regarding the wavelengths beyond the visible spectrum and have a wide range of applications including medical diagnosis \cite{2014Medical,2001Multimodal,1998Hyperspectral}, geothermal exploration \cite{2004Operational,2006Geothermal,2013Exploration}, agriculture \cite{2004Automated,2013A,2014The}. Unfortunately, different types of noise including stripes, deadlines, impulse noise, and Gaussian noise will be inevitably introduced in the hyperspectral imaging process, which considerably damages the image quality and limits further applications including image classification \cite{2005Kernel,2005Classification} and target detection \cite{2001Hyperspectral,2001Unsupervised}. Therefore, image denoising, as a preprocessing step in many computer vision tasks, is an essential and significant research topic.

During the past few decades, researches on hyperspectral images denoising has extensively proved that spatial nonlocal self-similarity and global spectral correlation are important prior information. A derivative-domain wavelet transform model is designed by Othman and Qian \cite{2006Noise} resorted to the signal regular dissimilarity along spatial and the spectral domain, which made an initial attempt to consider both priors. Subsequently, numerous denoising methods come up to benefit from the spatial and spectral features of HSIs. For example, a sparse representation model was developed by Qian et al. \cite{2013Hyperspectral} to utilize the two priors of HSI. Under a unified probabilistic framework, Zhong and Wang \cite{2013Multiple} considered spatial and spectral correlations and designed a multiple spectral-band conditional random field (MSB-CRF) method. A total variation model \cite{2012Hyperspectral} is proposed by Yuan et al. which adaptively adjust the denoising ability concerning the spatial property and noise intensity in each spectral band. To sum up, many advanced denoising techniques employed spectral and spatial information of the HSIs, including wavelet shrinkage \cite{2006Noise,2009Signal,2014Wavelet}, sparse representation \cite{2013Hyperspectral,DBLP:journals/tgrs/LuLFMB16}, etc.

For hyperspectral images, spectrally adjacent bands typically exhibit strong correlations due to the similar exposure time and wavelength, which reveals the low-rank property of hyperspectral images in the spectral domain. To explore the spectral low rankness, Zhang et al. \cite{2014Hyperspectral} split HSI into many overlapping full-band 3D cubes, then unfolded the 3D cubes to a matrix along with the spectral mode, and denoised under a low-rank matrix recovery (LRMR) \cite{DBLP:journals/corr/abs-0912-3599} framework. After that, they proposed a low-rank matrix approximation (LRMA) model \cite{2015Hyperspectral} to accommodate the differences in noise intensity at different bands which significantly enhance the denoising performance of LRMR. Based on LRMA, other approximation or relaxation of matrix rank were also proposed and integrated into the denoising model, such as $\gamma$-norm \cite{2017Denoising} and weighted Schatten $p$-norm \cite{2016Hyperspectral} etc. Follow this line, W. He et al. \cite{2015Total} employed low-rank matrix factorization (LRMF) to capture the spectral correlations as well as reduce the algorithm complexity. These low-rank-based methods have achieved comparable results for exploring the global spectral correlation, however, the solving process always containing singular value decomposition, which brings heavy computational burden when operated on high-dimensional data.

The spatial nonlocal self-similarity means that for any image patch extracted from a hyperspectral image, several patches similar to the reference patch can be found in the HSI, which suggests that the matrix generated by vectorizing each band of grouped similar patches as the column has a strong correlation can and is low-rank. Wang et al. \cite{2016Denoising} grouped the similar patches and explored the nonlocal self-similarity on a reconstructed group. Later, Xue et al. \cite{2018Joint} performed a low-rank regularized method on the similar patches in each cluster to characterize the spatial structure. Common operations of the above low-rank matrix-based methods are constructing a low-rank component according to clean HSI priors and explore its low-dimensional structure, which destroys the intrinsic structure of 3-D cube/image and has room for further improvement. Besides, the block matching process on high-dimensional data is a time-consuming operation.

In order to characterize the intrinsic structure of HSIs, many methods employed a low-rank tensor-based model and obtain better denoising performance. Tucker decomposition and CANDECOMP/PARAFAC (CP) decomposition are two typical tensor decomposition methods. Based on CP decomposition, Liu et al. \cite{2012Denoising} used parallel factor analysis (PARAFAC) model for HSI denoising and estimated the optimal rank of PARAFAC by calculating the covariance matrix after unfolding the matrix along $n$-mode. A new criterion is defined by Guo et al. \cite{2013Hyperspectral} to select the optimal CP rank, which calculates the minimum number of rank-1 tensors needed to describe a tensor. However, the prior information has not been fully taken into account as these methods concern HSI as a whole and ignore the information in other data structures, which usually results in suboptimal denoising performance. Based on Tucker decomposition, the ISTReg \cite{2016Multispectral} proposed a tensor sparsity measure composed of the core tensor and rank of matrix unfolded along with three modes, which makes the large computation problem more serious. Chang et al. \cite{2017Weighted} designed a weighted low-rank tensor recovery model applied on denoising and other visual tasks, where the singular values obtained by Tucker decomposition are treated differently concerning its contribution to rebuilding the original image. By formulating the denoising process as a low-rank tensor recovery (LRTR) problem, Fan et al. introduced a new tensor singular value decomposition (t-SVD) method \cite{Haiyan2017Hyperspectral}. \cite{0Hyperspectral} and \cite{2018Spatial} combined tucker decomposition and total variation method to describe the global correlation and local similarity within a patch. It is worth mentioning that Chang et al. \cite{2017Hyper} analyzed the low-rank characteristics in matrices and tensors and explained why the tensor-based method is better than the matrix-based method. Therefore, the tensor-based denoising method is an effective way to make full use of the global spectral correlation (GSC), nonlocal self-similarity (NSS) over space as well as maintaining the inherent structure, while at the expense of high complexity.

As for reducing the computational complexity, subspace-based \cite{Bioucas2008Hyperspectral} HSI denoising methods are effective techniques, which projects HSIs to eigen images based on high correlation existing among spectrum. Zhuang and Bioucas-Dias \cite{2018Hyperspectral} made a first try to explore the global spectral correlation by subspace representation, which reconstructed similar 3D patches from the representation coefficients image to a low-rank tensor and employed low-rank tensor factorization to remove noise. After that, also in the framework of subspace representation, they proposed a fast hyperspectral denoising (FastHyDe) algorithm for Gaussian and Poissonian noise \cite{2018Fastz}. Sun et al. \cite{2018Fasts} made the first attempt to remove HSI mixed noise based on subspace representation, and the superpixel segmentation is employed to exploit the spatial low rankness. He et al. \cite{2019Non} presented an integrated paradigm based on subspace representation and introduced an iterative mechanism on subspace. In addition, Cao et al \cite{DBLP:journals/staeors/CaoYZHXG19} introduced nonlocal low-rank factorization on spatial domain of the subspace. In the above methods, the low rankness of the spectrum is characterized by subspace representation, and the subspace-based denoising method is capable of reducing complexity significantly.

As we know, real-world HSIs usually disturbed by Gaussian noise and different types of sparse noise due to the uncertainties during image acquisition process. Motivated by \cite{2017Weighted} and \cite{DBLP:journals/staeors/CaoYZHXG19}, we propose a method to remove mixed noise in hyperspectral image by combining subspace representation with weighted low-rank tensor regularization. On the one hand, the subspace representation allows the method to impose the spectral correlation and convert the problem into an estimate of the subspace coefficients, thus simplifying the calculation. On the other hand, the introduced weighted low-rank tensor regularization term characterize the priors in the reduced image effectively, where the singular value of the core tensor is punished according to its importance for recovering real HSI. The contributions of our work are summarized as follow:
\begin{enumerate} 
 \setlength{\itemsep}{-2ex}
 \setlength{\parskip}{0ex}
 \setlength{\parsep}{0ex}
\item We designed a novel mixed noise removal model for hyperspectral images. Based on the high spectral correlation, the subspace representation decomposes the noisy hyperspectral image into a mode-3 tensor-matrix product, which simplifies the estimation of clean images. We reconstruct the similar patches groups searched from reduced image to obtain a tensor with low rankness, and introduce a weighted low-rank tensor regularization term to constrain the low-rank characteristics considering the physical meaning of core tensor. In addition, the sparse noise constrained by ${l_1}$-norm is also integrated into the model to remove the sparse noise. In the denoising process, the intrinsic structure of the tensor is better preserved and the computational complexity is lower than comparison algorithms.\hfil\break
\item An alternating minimization-based algorithm is designed to solve the proposed denoising model and it is possible to obtain an optimal solution.\hfil\break
\item Furthermore, simulated and real datasets experiments prove that our method is superior to existing methods both quantitatively and qualitatively.\hfil\break
\end{enumerate}

\section{Notations and Preliminaries}
In this article, the tensor is denoted by capitalized calligraphic letter, i.e. $\mathcal{X}$;  matrix by uppercase letter, i.e. $X$; vector by lowercase letter, i.e. $x$. The $i$-th entry of vectors $x$ is represented as ${x_i}$, the element in the $i$-th row and $j$-th column of the matrices $X$ is expressed as ${X_{ij}}$, the element $(i,j,k)$ of tensors $\mathcal{X}$ is expressed as ${x_{ijk}}$. The Frobenius norm of a $N$ order tensor $\mathcal{X} \in {\mathbb{R}^{{I_1} \times {I_2} \times  \cdots  \times {I_N}}}$ is defined as ${\left\| \mathcal{X} \right\|_F} = {\left( {\sum\limits_{{i_1} = 1}^{{I_1}} {\sum\limits_{{i_2} = 1}^{{I_2}}  \cdots  } \sum\limits_{{i_N} = 1}^{{I_N}} {{{\left| {{x_{{i_1}{i_2} \cdots {i_N}}}} \right|}^2}} } \right)^{1/2}}$, its ${l_1}$ norm is 
defined as ${\left\| \mathcal{X} \right\|_1} = \sum\limits_{{i_1} = 1}^{{I_1}} {\sum\limits_{{i_2} = 1}^{{I_2}}  \cdots  } \sum\limits_{{i_N} = 1}^{{I_N}} {\left| {{x_{{i_1}{i_2} \cdots {i_N}}}} \right|} $. The mode-$n$ vector of a $N$-order tensor $\mathcal{X} \in {\mathbb{R}^{{I_1} \times {I_2} \times  \cdots  \times {I_N}}}$ is an ${I_n}$ dimensional vector with ${i_n}$ as the element subscript variable, and all other subscripts $\left\{ {{i_1}, \cdots ,{i_N}} \right\}\backslash {i_n}$ are fixed, denoted as ${X_{{i_1} \cdots {i_{n - 1}}:{i_{n + 1}} \cdots {i_N}}}$. The matrixization of a tensor refers to the transformation of reorganizing an $N$-order tensor into a matrix form, including horizontal expansion and vertical expansion. The mode-$n$ vectors of the $N$-order tensor $\mathcal{X} \in {\mathbb{R}^{{I_1} \times {I_2} \times  \cdots  \times {I_N}}}$ arranged in the horizontal direction called mode-$n$ unfolding along the horizontal, expressed as ${X_{(n)}}$, ${X_{(n)}} \in {\mathbb{R}^{\left( {{I_1} \cdots {I_{n - 1}}{I_{n + 1}} \cdots {I_N}} \right) \times {I_n}}}$. The mode-$n$ vectors of the tensor arranged in the vertical direction called mode-$n$ unfolding along the vertical, denoted as ${{X}^{(n)}}$, ${{X}^{(n)}} \in {\mathbb{R}^{\left( {{I_1} \cdots {I_{n - 1}}{I_{n + 1}} \cdots {I_N}} \right) \times {I_n}}}$. The mode-$n$ tensor-matrix product of a $N$-order tensor $\mathcal{X} \in {\mathbb{R}^{{I_1} \times {I_2} \times  \cdots  \times {I_N}}}$ and a ${J_n} \times {I_n}$ matrix ${U}$ is represented by $\mathcal{X}{ \times _n}{U}$, which is a ${I_1} \cdots {I_{n - 1}} \times {J_n} \times {I_{n + 1}} \cdots {I_N}$ tensor, its elements are defined as ${\left( {\mathcal{X}{ \times _n}{U}} \right)_{{i_1} \cdots {i_{n - 1}}j{i_{n + 1}} \cdots {i_N}}}\mathop  = \limits^{{\text{ }}def{\text{ }}} \sum\limits_{{i_n} = 1}^{{I_n}} {{x_{{i_1}{i_2} \cdots {i_N}}}} {u_{j{i_n}}}$, $j = 1, \cdots ,{J_n}$, ${i_k} = 1, \cdots ,{I_k}$, $k = 1, \cdots ,N$. Tucker decomposition is a kind of high-order singular value decomposition that each ${I_1} \times {I_2} \times  \cdots  \times {I_N}$ tensor $\mathcal{X}$ can be decomposed into a mode-$n$ product $\mathcal{X} = \mathcal{G}{ \times _1}{U^{(1)}}{ \times _2}{U^{(2)}} \cdots { \times _N}{U^{(N)}}$, where ${U^{(n)}}$ is a semi-orthogonal ${I_n} \times {J_n}$ matrix, the core tensor $\mathcal{G} \in \mathbb{R}^{J_{1} \times J_{2} \times \cdots \times J_{N}}$ is generally a full tensor, representing the interaction among each modes of matrices ${U^{(n)}},n = 1,...,N$. Besides, the kronecker product of $m \times n$ matrix ${A}=\left[{a}_{1}, \cdots, {a}_{n}\right]$ and $p \times q$ matrix $B$ is denoted as ${A} \otimes {B}=\left[{a}_{1} {B}, \cdots, {a}_{n} {B}\right]=\left[a_{i j} {B}\right]_{i=1, j=1}^{m, n}$, which is a matrix of size $m p \times n q$.

\section{ Problem formulation }
\subsection{ Problem formulation }
The noisy hyperspectral image $\mathcal{Y} \in {\mathbb{R}^{{n_1} \times {n_2} \times {n_3}}}$ is assumed to be the sum of clean signal $\mathcal{X}$ and noise, where ${n_1}$, ${n_2}$ are the length and width of hyperspectral image, and ${n_3}$ is the number of spectral bands. The noise is modeled as random noise $\mathcal{N}$ and sparse noise $\mathcal{S}$ according to the characteristics of the noise intensity distribution. Therefore, the degradation model of removing HSI mixed noise can be expressed as
\begin{eqnarray}
\mathcal{Y} = \mathcal{X} + \mathcal{N} + \mathcal{S}
\end{eqnarray}
The purpose of hyperspectral image denoising is to recover clean signal $\mathcal{X}$ from the observed image $\mathcal{Y}$, which is a difficult ill-posed problem. The key to solving this problem is to establish appropriate regularization based on reasonable HSI prior. On this basis, the denoising model of the hyperspectral image can be expressed as
\begin{eqnarray}
\{\hat{\mathcal{X}}, \hat{\mathcal{S}}\}=\mathop {\arg \min }\limits_{\mathcal{X},\mathcal{S}} \frac{1}{2}\left\| {\mathcal{Y} - \mathcal{X} - \mathcal{S}} \right\|_F^2 + {\lambda _1}{J_1}(\mathcal{X}) + {\lambda _2}{J_2}(\mathcal{S})
\end{eqnarray}
where $\|\cdot\|_{F}^{2}$ denotes the Frobenius norm of a tensor and measures the error bounds between the reconstructed image and noisy HSI. ${J_1}(\mathcal{X})$ and ${J_2}(\mathcal{S})$ are the regularization terms of the clean image $\mathcal{X}$ and sparse noise $\mathcal{S}$, respectively. ${\lambda _1}$ and ${\lambda _2}$ are the parameters to balance the data fidelity term and the regularization term. Here, the sparsity of $\mathcal{S}$ is constraint by ${l_1}$ norm. So (2) can be converted to:
\begin{eqnarray}
\{\hat{\mathcal{X}}, \hat{\mathcal{S}}\}=\mathop {arg\min }\limits_{\mathcal{X},\mathcal{S}}\frac{1}{2}\|\mathcal{Y}-\mathcal{X}-\mathcal{S}\|_{F}^{2}+\lambda_{1} J_{1}(\mathcal{X})+\lambda_{2}\|\mathcal{S}\|_{1}
\end{eqnarray}

\subsection{ Subspace representation }
Due to the high correlation in spectral, the combination of a few pure endmembers can express any spectral signature linearly. Assuming that HSI $\mathcal{X}$ is located in a k-dimensional spectral subspace ${\mathcal{H}_k}$, the clean hyperspectral image $\mathcal{X}$ can be expressed as $\mathcal{X} = \mathcal{Z}{ \times _3}A$, where $A \in {\mathbb{R}^{{n_3} \times k}}$ is an orthogonal matrix holding the basis of the reduced image $\mathcal{Z} \in {\mathbb{R}^{{n_1} \times {n_2} \times k}}$. Therefore, HSI denoising model (3) can be reformulated as follows:
\begin{eqnarray}
\begin{array}{l}
\begin{gathered}
  \{\hat{A}, \hat{\mathcal{Z}},\hat{\mathcal{S}}\}  = \mathop {arg\min }\limits_{\mathcal{X},\mathcal{S}} \frac{1}{2}\left\| {\mathcal{Y} - \mathcal{Z}{ \times _{\text{3}}}A - \mathcal{S}} \right\|_F^2 + {\lambda _1}{J_1}(\mathcal{Z}) \\
  \quad\quad\quad\quad\quad\quad\quad\quad\quad + {\lambda _2}{\left\| \mathcal{S} \right\|_1} \hfill 
  s.t.A{A^T} = {I_k} \hfill \\ 
\end{gathered}
\end{array}
\end{eqnarray}
W. He et al. \cite{8954159} has proved that the orthogonal constraint help to exclude noise from $A$ and maintain the distribution of noise in the reduced image. Besides, the benefits of factorizing the HSI to a mode-3 tensor-matrix product include preserving the integrity of the structural information and reducing the computational cost significantly. 

\subsection{ Weighted low-rank tensor regularization }
In the last decades, nonlocal similar patch-based methods have demonstrated that grouping similar patches do enhance the low-rank nature. Therefore, we exploit the two inherent priors of the reduced image to obtain a tensor with a more significant low-rank property by aggregating the similar full-band patches, which benefits the HSI image denoising. In practical applications, 3-D patches similar to reference 3-D patches are searched in adjacent areas. Define that patch size is $p \times p \times k$, and the number of similar patches is $q$. The process of constructing low-rank tensor ${\mathcal{Z}_i}$ from $\mathcal{Z}$ is as follows:
\begin{enumerate}
 \setlength{\itemsep}{-2ex}  
 \setlength{\parskip}{0ex} 
 \setlength{\parsep}{0ex}
\item Searching: patches are sampled in the neighborhood, and most of the pixels appeared in several patches.\hfil\break
\item Block matching: calculate the euclidean distance between the reference image patch and the selected patches, then retain the first $q$ patches that more similar to the reference image.\hfil\break
\item Unfolding: reshape each 3-D image patch in the group into a ${p^{\text{2}}} \times k$ matrix to get the mode-3 vectors of these patches.\hfil\break
\item Stacking: stack the $q$ mode-3 vectors obtained in step 3) into a tensor of size ${p^{\text{2}}} \times q \times k$, denoted as ${\mathcal{Z}_i}$.\hfil
\end{enumerate}
The above process is described as ${\mathcal{Z}_i}{\text{ = }}{\Re _i}\mathcal{Z}$, where $i$ is the top-left corner pixel coordinates, and ${\Re _i}$ represents a series of operations extracting the low-rank tensor ${\mathcal{Z}_i}$ from the reduced image $\mathcal{Z}$.

\begin{figure*}
\centering
\includegraphics[width=0.8\textwidth]{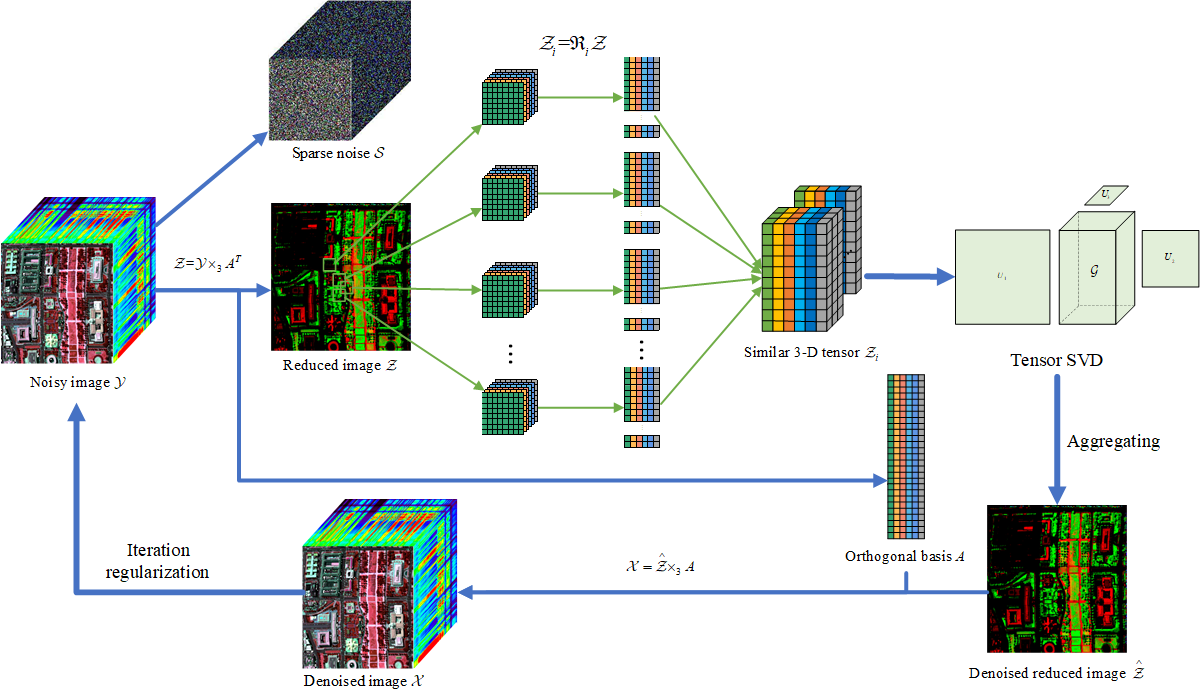}
\caption{ The denoising process of proposed method. It includes subspace representation $\mathcal{Z}=\mathcal{Y} \times_{\mathrm{y}} A^{T}$, block matching $\mathcal{Z}_{i}=\mathfrak{R}_{i} \mathcal{Z}$, Tensor SVD and iteration regularization, etc. }
\label{Fig.1}
\end{figure*}

The different performances of \cite{Haiyan2017Hyperspectral,0Hyperspectral,2018Spatial} are caused by the different regularization term used for the low-rank tensor, falling into this line, we introduce a low-rank tensor regularization term \cite{2017Weighted} to exploit the prior knowledge of ${\mathcal{Z}_i}$ while preserving intrinsic structure. Therefore, estimating the corresponding low-rank approximation, i.e. clean tensor ${{\mathcal{L}_i}}$ from ${\mathcal{Z}_i}$ can be written as:
\begin{eqnarray}
\mathop { \hat{\mathcal{L}_i}} = \mathop {\arg \min }\limits_{{\mathcal{L}_i}} \frac{1}{{\sigma _i^2}}\left\| {{\mathcal{Z}_i} - {\mathcal{L}_i}} \right\|_F^2 + {\left\| {{\mathcal{L}_i}} \right\|_{w,*}}
\end{eqnarray}
where ${\left\| {{\mathcal{L}_i}} \right\|_{w,*}} = \sum\limits_j {{{\left| {{w_j}{\sigma _j}\left( {{\mathcal{L}_i}} \right)} \right|}_1}} , w = \left[ {{w_1}, \ldots ,{w_n}} \right]$, ${w_j}$ are the non-negative weights assigned to ${\sigma _j}\left( {{\mathcal{L}_i}} \right)$, and $\sigma _i^2$ is the noise variance.

In general, the singular values in the core tensors are mostly close to zero due to the strong correlation among ${\mathcal{Z}_i}$. Large singular values represent the energy of the major components of ${\mathcal{Z}_i}$, while small singular values tend to be larger than clean tensor due to noise interference. Therefore, it is rational to impose less penalty for larger singular values and more penalty for smaller singular values. A natural approach is that the value of weight assigned to ${{\sigma _j}\left( {{\mathcal{L}_i}} \right)}$ is inversely proportional to ${{\mathcal{L}_i}}$. According to WLRTR \cite{2017Weighted}, we set
\begin{eqnarray}
{w_j} = c\sqrt q /\left( {\left| {{\sigma _j}\left( {{\mathcal{L}_i}} \right)} \right| + \varepsilon } \right)
\end{eqnarray}
where $\varepsilon $ is a small constant that avoids being divided by zero, and $c > 0$ is a constant.

\section{ Denoising model and optimization algorithm }
\subsection{ Denoising model and optimization algorithm }
By integrating the weighted low-rank tensor regularization with the framework of subspace representation, the HSI mixed noise removal model is proposed as
\begin{eqnarray}
\begin{array}{l}
  \{ \hat{A}, \hat{\mathcal{Z}}, \hat{\mathcal{L}_i}, \hat{\mathcal{S}}  \}  = \mathop {\arg \min }\limits_{A,\mathcal{Z},{\mathcal{L}_i},\mathcal{S}} \frac{1}{2}\left\| {\mathcal{Y} - \mathcal{Z}{ \times _3}A - \mathcal{S}} \right\|_F^2 \\ 
  \quad\quad\quad + {\lambda _1}\mathop \sum \limits_i (\frac{1}{{\sigma _i^2}}\left\| {{\Re _i}\mathcal{Z} - {\mathcal{L}_i}} \right\|_F^2 + {{\left\| {{\mathcal{L}_i}} \right\|}_{w,*}}) \\
  \quad\quad\quad + {\lambda _2}{{\left\| \mathcal{S} \right\|}_1}\quad {\text{ s}}{\text{.t}}{\text{.  }}A{A^T} = {I_k} 
\end{array}
\end{eqnarray}
We introduce an algorithm based on an alternating minimization strategy to optimize the proposed model efficiently. The solution procedure of the model (7) mainly includes two steps: solving the low-rank tensor regularization term and the reduced image ${\mathcal{Z}}$. 

\subsubsection{low-rank tensor regularization term}
In this subproblem, we need to estimate ${\mathcal{L}_i}$ from the low-rank tensor ${\mathcal{Z}_i}$. By ignoring the variables irrelevant to ${\mathcal{L}_i}$ in (7), we can get the subproblem:
\begin{eqnarray}
\mathop { \hat{\mathcal{L}_i}}  =\mathop {\arg \min }\limits_{{\mathcal{L}_i}} \frac{1}{{\sigma _i^2}}\left\| {{\mathcal{Z}_i} - {\mathcal{L}_i}} \right\|_F^2 + {\left\| {{\mathcal{L}_i}} \right\|_{w,*}}
\end{eqnarray}
\cite{2017Weighted} developed an alternating direction minimization method for solving the problem (8). By replacing ${\mathcal{L}_i}$ with the corresponding $Tucker$ decomposition, the optimization problem shown in (8) is equivalent to the following problem:
\begin{eqnarray}
\begin{array}{l}
\begin{gathered}
  \{ \hat{\mathcal{G}_i},\hat{{U}_1},\hat{{U}_2},\hat{{U}_3} \}  = \mathop {\arg \min }\limits_{{\mathcal{G}_i},{{\mathbf{U}}_1},{{\mathbf{U}}_2},{{\mathbf{U}}_3}} \left\| {{\mathcal{Z}_i} - {\mathcal{G}_i}{ \times _1}{U_1}{ \times _2}{U_2}{ \times _3}{U_3}} \right\|_F^2 \\
  \quad\quad\quad\quad\quad + \sigma _i^2{\left\| {{w_i}^\circ {\mathcal{G}_i}} \right\|_1} \hfill {\text{s}}{\text{.t}}{\text{. }}U_j^T{U_j} = I,j = 1,2,3 \hfill \\ 
\end{gathered}
\end{array}
\end{eqnarray}
where $j$ represents the mode index of a 3-order tensor, and $ \circ $ denotes the element-wise multiplication. To estimate ${U_1},{U_2},{U_3}$ iteratively, let ${W_1} = {\mathcal{Z}_i}_{(1)}\left( {{U_3} \otimes {U_2}} \right)$, where $\otimes$ denotes the Kronecker product, then 
\begin{eqnarray}
{U_1} = P{Q^T}
\end{eqnarray}
where ${W_1} = P\Sigma {Q^T}$ is the singular value decomposition of ${W_1}$. By assuming ${W_2} = {\mathcal{Z}_i}_{(2)}\left( {{U_3} \otimes {U_1}} \right)$, ${W_3} = {\mathcal{Z}_i}_{(3)}\left( {{U_2} \otimes {U_1}} \right)$, ${U_2}$ and ${U_3}$ can be optimized by the similar procedures, respectively. Let ${\mathcal{O}_i} = {\mathcal{Z}_i}{ \times _1}U_1^T{ \times _2}U_2^T{ \times _3}U_3^T$, ${\mathcal{G}_i}$ can be obtained through
\begin{eqnarray}
{\mathcal{G}_i} = \operatorname{sign} ({\mathcal{O}_{\text{i}}})\max (|{\mathcal{O}_{\text{i}}}| - {w_{\text{i}}}\sigma _{\text{i}}^2/2,0)
\end{eqnarray}
After obtained four variables, the mode-3 tensor-matrix multiplication is performed to restored optimal ${\mathcal{L}_i}$.

\renewcommand{\algorithmicrequire}{\textbf{Input:}}  
\renewcommand{\algorithmicensure}{\textbf{Output:}} 

\begin{algorithm}
  \caption{ SWLRTR for hyperspectral image denoising } 
  \label{alg::conjugateGradient}
  \begin{algorithmic}[1]
    \Require
      $\mathcal{Y}$: noisy hyperspectral image;
      $k$: subspace dimension;
      $p$: image patch size;
      $q$: number of non-local similar patches;
    \Ensure
      Denoised image $\mathcal{X}$
    \State Initialize: ${\mathcal{X}^1} = {\mathcal{Y}^1} = \mathcal{Y}$, $\mathcal{S} = 0$;
    \For{$n = 1,2,3...N$ }
      \State Estimate orthogonal basis matrix ${A^n}$ by Hysime \cite{Bioucas2008Hyperspectral};
      \State Compute the reduced image ${\mathcal{Z}^n}$ according to ${\mathcal{Z}^n} = {\mathcal{Y}^n}{ \times _3}{A^n}$;
      \For{each patch $\mathcal{Z}_i^n$ in ${\mathcal{Z}^n}$ }
        \State Find similar 3-D patch group to form tensor ${\mathcal{Z}_i}$;
        \For { (low-rank tensor recovery) ${n_0} = 1...{N_0}$ }
        \State Estimate weight vector $w$ via Eq.(6);
        \State Estimate ${U_j}$ via Eq.(10);
        \State Estimate the core tensor ${\mathcal{G}_i}$ via Eq.(11);
        \State Get the Estimation $\hat {{\mathcal{L}_i}}  = {\mathcal{G}_i}{ \times _1}{U_1}{ \times _2}{U_2}{ \times _3}{U_3}$;
        \EndFor
      \EndFor
      \State Aggregate $\hat {{\mathcal{L}_i}}$ to form $\hat{{\mathcal{Z}^n}}$ ;
      \Repeat
        \State update $\mathcal{S}$ via Eq. (14);
        \State update ${\mathcal{Z}^n}$ via Eq.(16);
        \State update ${A^n}$ via Eq.(18);
      \Until{convergence}
      \State Compute denoised ${\mathcal{X}^n}$ according to ${\mathcal{X}^n} = {\mathcal{Z}^n}{ \times _{\text{3}}}{A^n}$;
      \State Iterative regularization ${\mathcal{Y}^{n + 1}} = \alpha {\mathcal{X}^n} + (1 - \alpha )\mathcal{Y}$ \\
      \quad\quad$k = k + \beta  \times n$
    \EndFor
  \end{algorithmic}
\end{algorithm}

\subsubsection{ the reduced image ${\mathcal{Z}}$ }
With ${\mathcal{L}_i}$ fixed, the formulation (7) is converted to the following minimization problem:
\begin{eqnarray}
\begin{array}{l}
\{\hat{A}, \hat{\mathcal{Z}}, \hat{\mathcal{S}}\}={\mathop {\arg \min }\limits_{A,\mathcal{Z},\mathcal{S}} }\frac{1}{2}\left\|\mathcal{Y}-\mathcal{Z} \times_{3} A-\mathcal{S}\right\|_{F}^{2}+\lambda_{2}\|\mathcal{S}\|_{1} \\
\quad \quad \quad \quad \quad +\lambda_{1} \sum \frac{1}{\sigma_{i}^{2}}\left\|\mathfrak{R}_{i} \mathcal{Z}-\mathcal{L}_{i}\right\|_{F}^{2} \text { s.t. } \quad A A^{T}=I_{k}
\end{array}
\end{eqnarray}
The augmented lagrangian function of the subproblem (12) is
\begin{eqnarray}
\begin{array}{l}
\{\hat{A}, \hat{\mathcal{Z}}, \hat{\mathcal{S}}\}={\mathop {\arg \min }\limits_{A,\mathcal{Z},\mathcal{P}} }\frac{1}{2}\left\|\mathcal{Y}-\mathcal{Z} \times_{3} A-\mathcal{S}\right\|_{F}^{2}+\lambda_{2}\|\mathcal{S}\|_{1} \\
\quad\quad\quad\quad\quad +\lambda_{1} \sum \frac{1}{\sigma_{i}^{2}}\left\|\Re_{i} \mathcal{Z}-\mathcal{L}_{i}\right\|_{F}^{2}+\xi_{\left\{I_{k}\right\}}\left(A A^{T}\right)
\end{array}
\end{eqnarray}
where ${\xi _{\{ {I_k}\} }}( \cdot )$ is the indicator function. To solve (13), we can alternately update $\mathcal{S}$, $\mathcal{Z}$, $A$:
\begin{itemize}
\item[$\bullet$] Update $\mathcal{S}$: By fixing $\mathcal{Z}$ and $A$, (13) can be written as
\begin{eqnarray}
\hat{\mathcal{S}} =\mathop {\arg \min }\limits_\mathcal{S}  \frac{1}{2}\left\|\mathcal{Y}-\mathcal{Z} \times_{3} A-\mathcal{S}\right\|_{F}^{2}+\lambda_{2}\|\mathcal{S}\|_{1}
\end{eqnarray}
Its closed-form solution is $S_{\lambda_{2}}\left(\mathcal{Y}-\mathcal{Z} \times_{3} A\right)$, where ${S_\omega }(x) = \operatorname{sgn} (x)\max (|x| - \omega ,0)$ is the soft threshold operation.
\item[$\bullet$] Update $\mathcal{Z}$: By fixing $\mathcal{S}$ and $A$, (13) can be written as
\begin{eqnarray}
\begin{array}{l}
\hat{\mathcal{Z}}=\underset{\mathcal{Z}}{\arg \min } \frac{1}{2}\left\|\mathcal{Y}-\mathcal{Z} \times_{3} A-\mathcal{S}\right\|_{F}^{2} \\
\quad \quad +\lambda_{1} \sum_{i} \frac{1}{\sigma_{i}^{2}}\left\|\mathfrak{R}_{i} \mathcal{Z}-\mathcal{L}_{i}\right\|_{F}^{2}
\end{array}
\end{eqnarray}
Equation (15) is a quadratic optimization problem and has a closed-form solution: 
\begin{eqnarray}
\begin{array}{l}
\hat{\mathcal{Z}}   = {\left( {{{\mathbf{I}}_k} + {\lambda _1}\sum\limits_i {\frac{2}{{\sigma _i^2}}} \Re _i^T{\Re _i}} \right)^{ - 1}}\\
\quad\quad \times \left( {(\mathcal{Y} - \mathcal{S}){ \times _{\text{3}}}{A^T} + {\lambda _1}\sum\limits_i {\frac{2}{{\sigma _i^2}}} \Re _i^T{\mathcal{L}_i}} \right)
\end{array}
\end{eqnarray}
\item[$\bullet$] Update $A$: By fixing $\mathcal{S}$ and $\mathcal{Z}$, (13) can be written as
\begin{eqnarray}
\hat{A}=\mathop {\arg \min }\limits_{A,A{A^T} = {I_k}} \frac{1}{2}\left\|\mathcal{Y}-\mathcal{Z} \times_{3} A-\mathcal{S}\right\|_{F}^{2}
\end{eqnarray}
Let $M = {(\mathcal{Y} - \mathcal{S})^{(3)}}{\mathcal{Z}_{(3)}}$, $M = U\Lambda {V^T}$ is assumed to be the singular value decomposition of $M$, then
\begin{eqnarray}
\hat{A} = U{V^T}
\end{eqnarray}
\end{itemize}

\begin{figure}
    \centering
    \subfigure[Gaussian+impulse]{
        \includegraphics[width=0.45\linewidth]{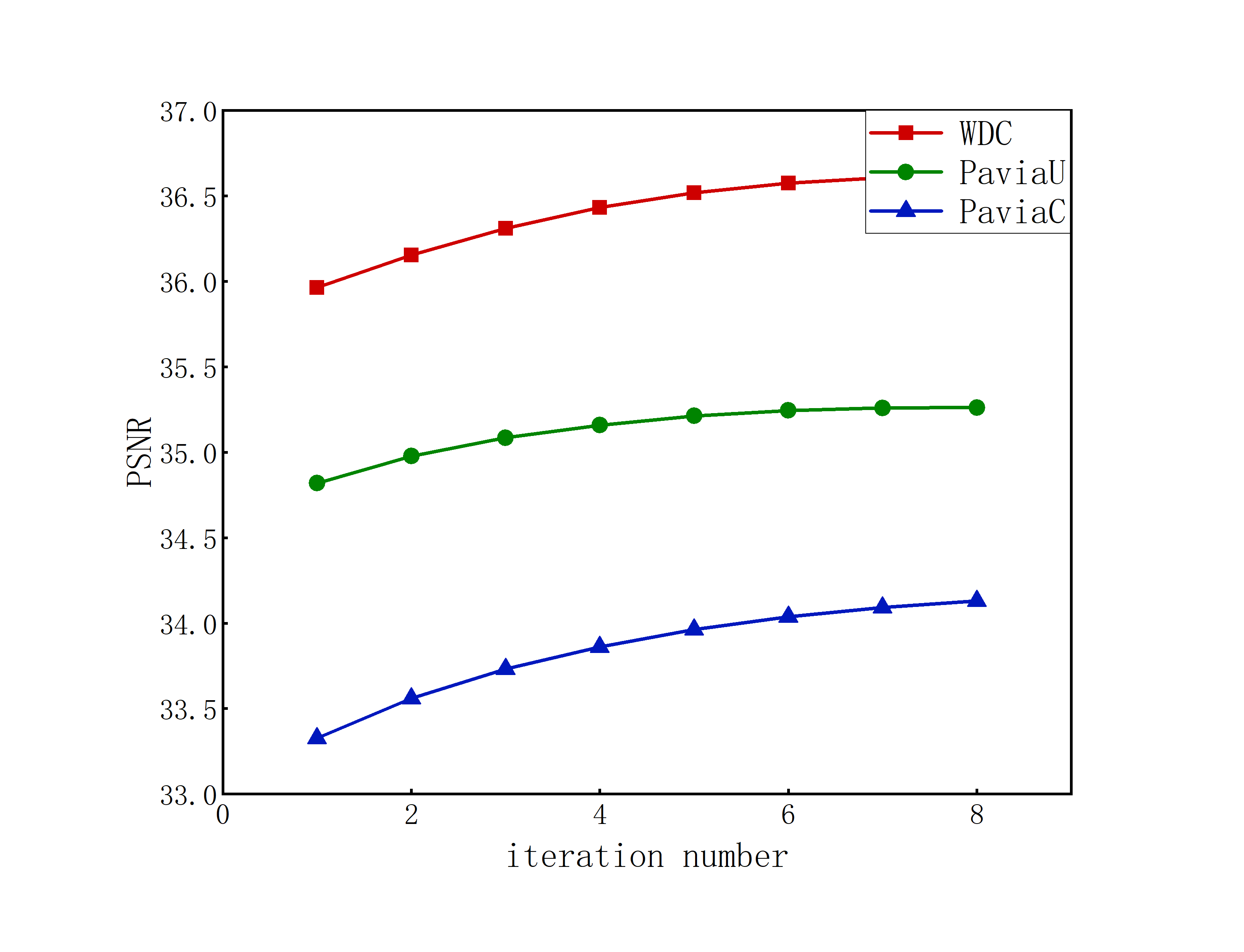}
    }
    \subfigure[Gaussian+impulse+deadlines]{
        \includegraphics[width=0.45\linewidth]{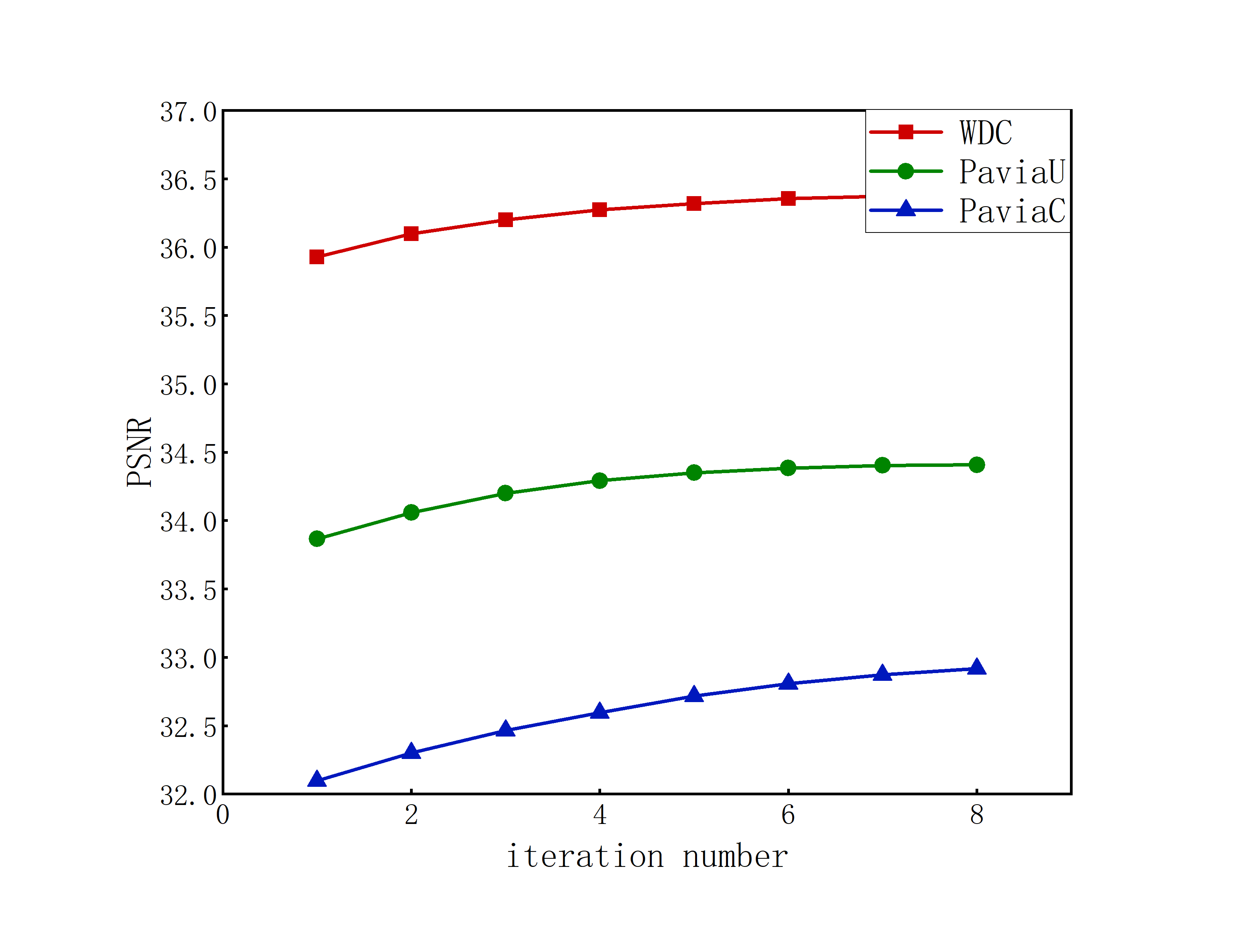}
    }
    \centering
    \caption{ PSNR values with the increase of iteration }
    \label{Fig.2}
\end{figure}

\begin{figure}
    \centering
    \subfigure[]{
        \includegraphics[width=0.29\linewidth]{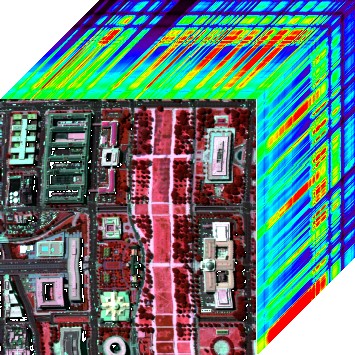}
    }
    \subfigure[]{
        \includegraphics[width=0.29\linewidth]{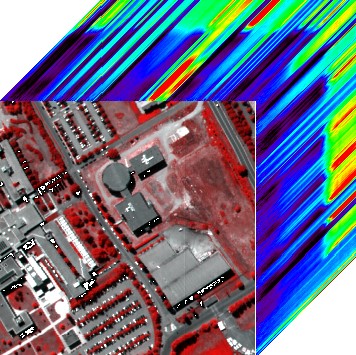}
    }
    \subfigure[]{
        \includegraphics[width=0.29\linewidth]{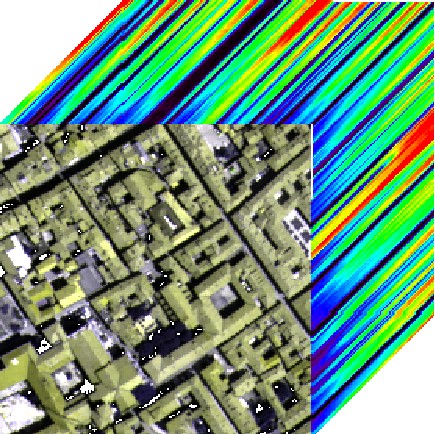}
    }
    \centering
    \caption{(a) Washington DC Mall dataset (R:121,G:30,B:49) (b) Pavia University dataset (R:75,G:58,B:60) (c) Pavia Center dataset (R:24,G:30,B:6)}
    \label{Fig.3}
\end{figure}

\begin{figure*}
    \centering
    \subfigure[]{
        \includegraphics[width=0.23\textwidth]{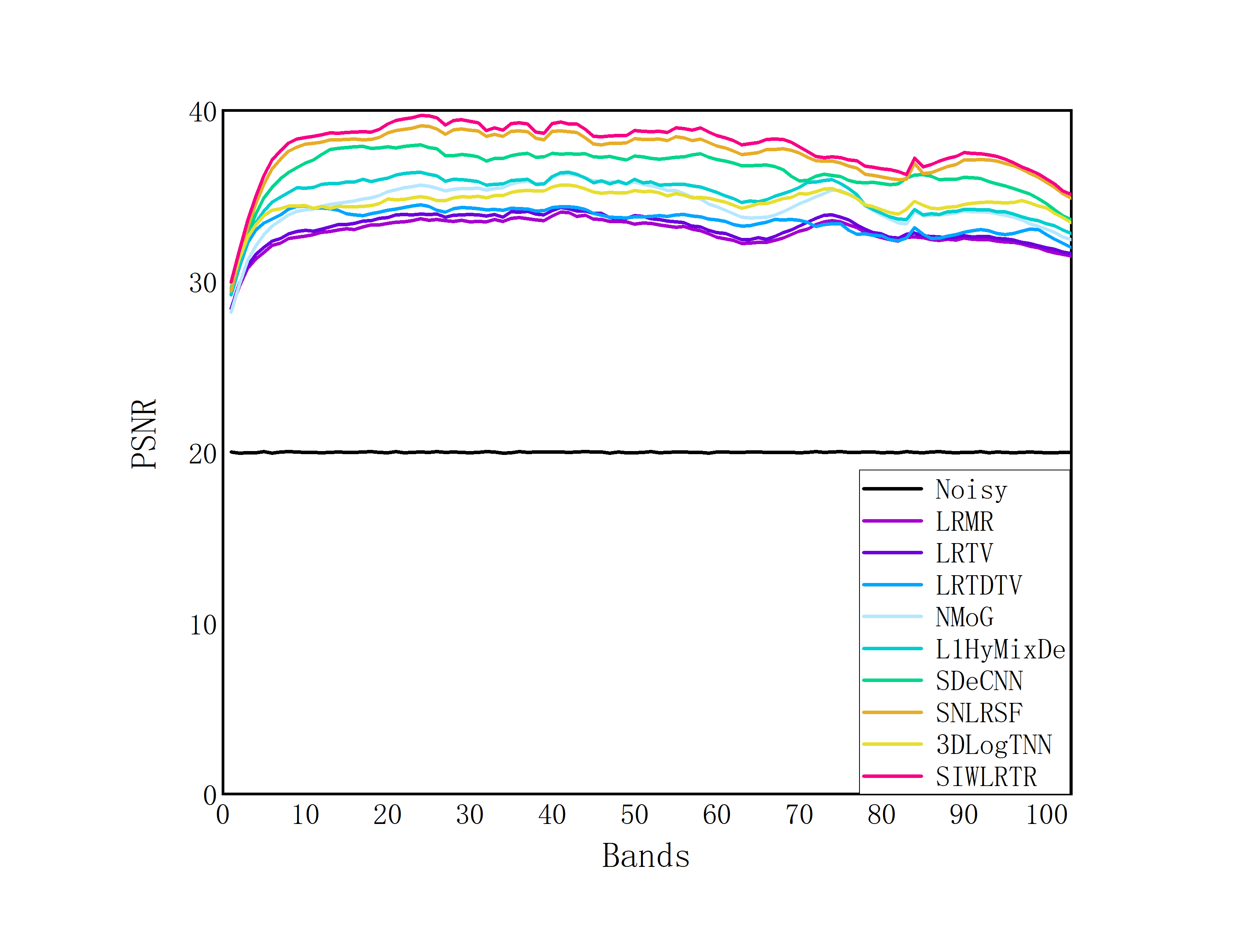}
    }
    \subfigure[]{
        \includegraphics[width=0.23\textwidth]{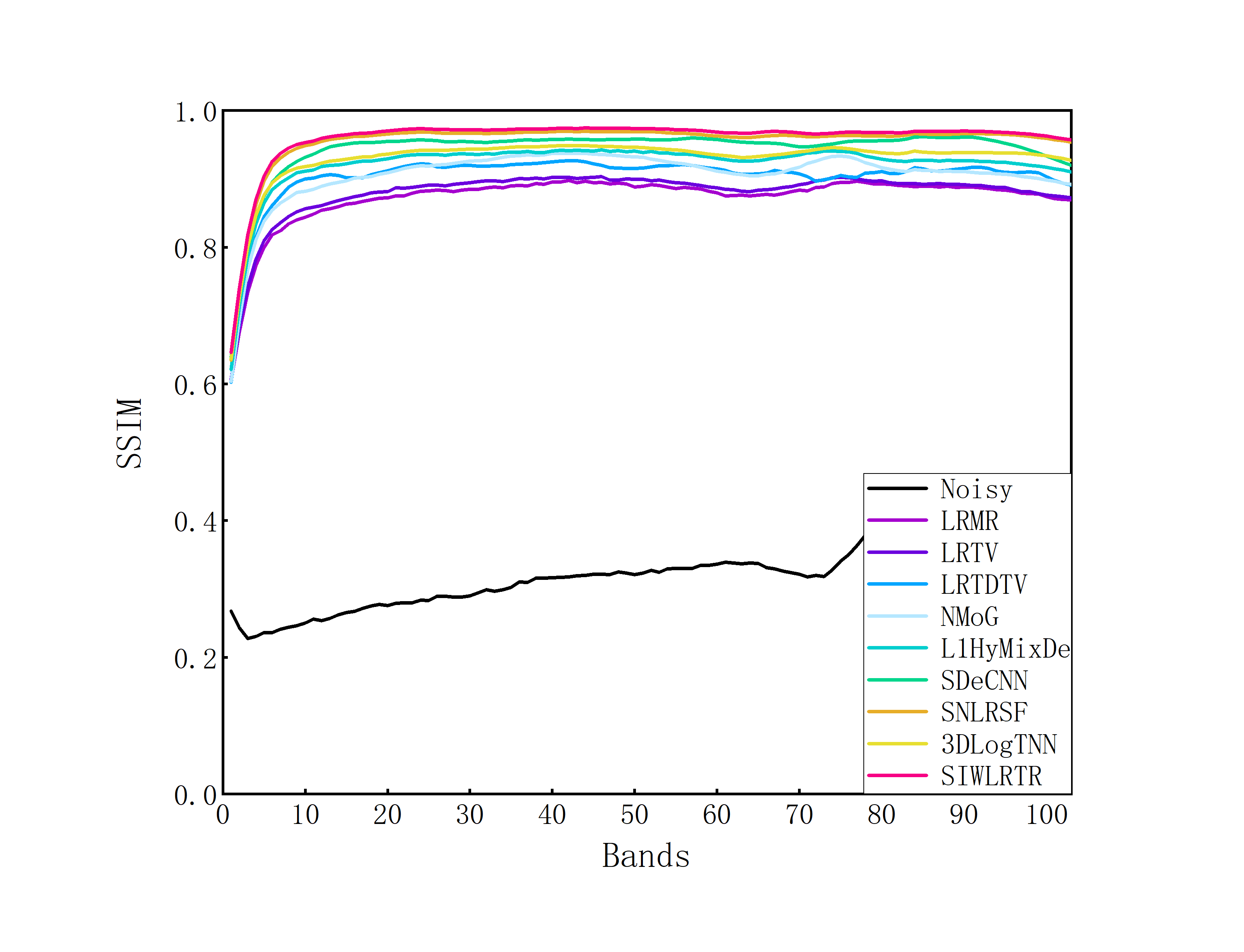}
    }
    \subfigure[]{
        \includegraphics[width=0.23\textwidth]{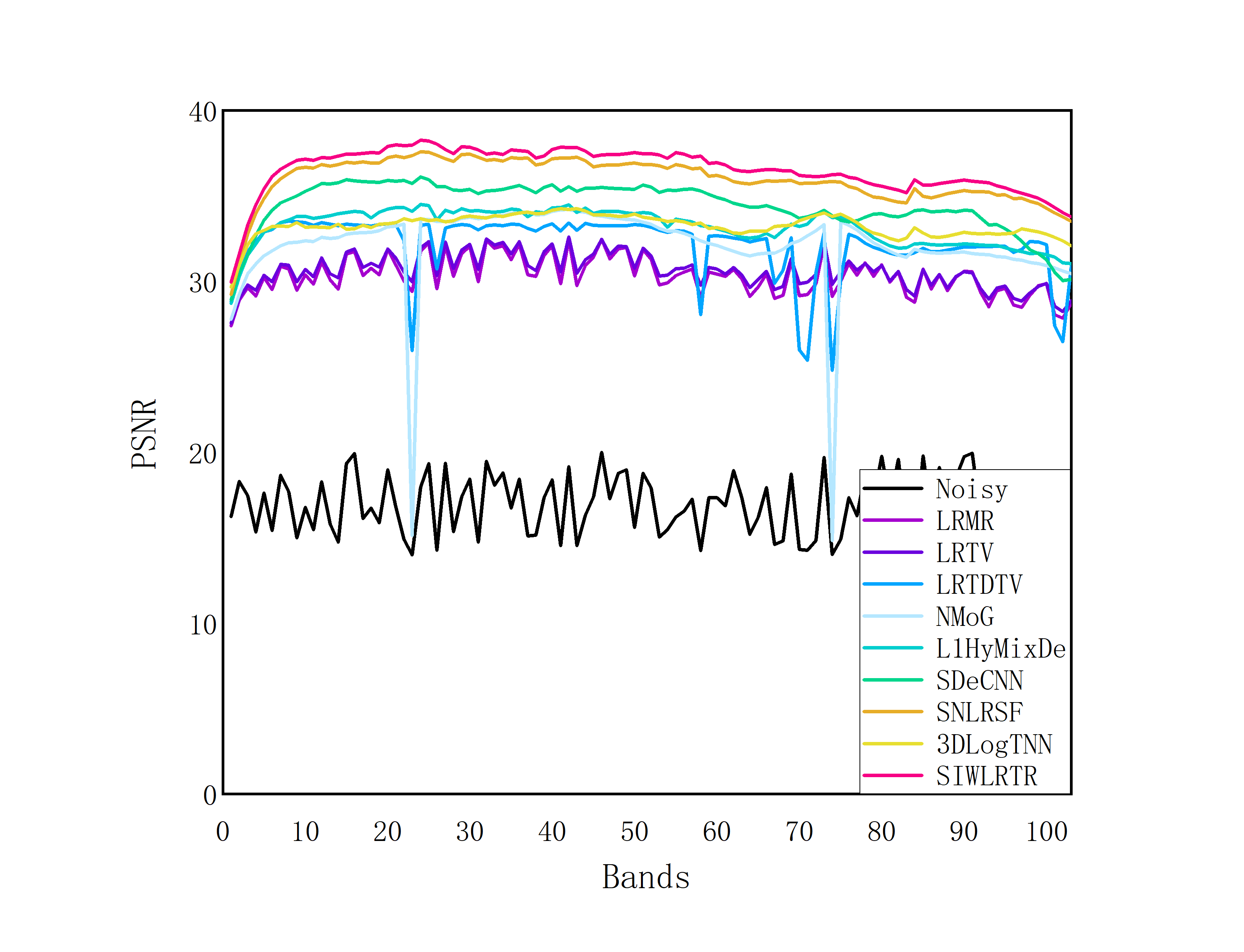}
    }
     \subfigure[]{
        \includegraphics[width=0.23\textwidth]{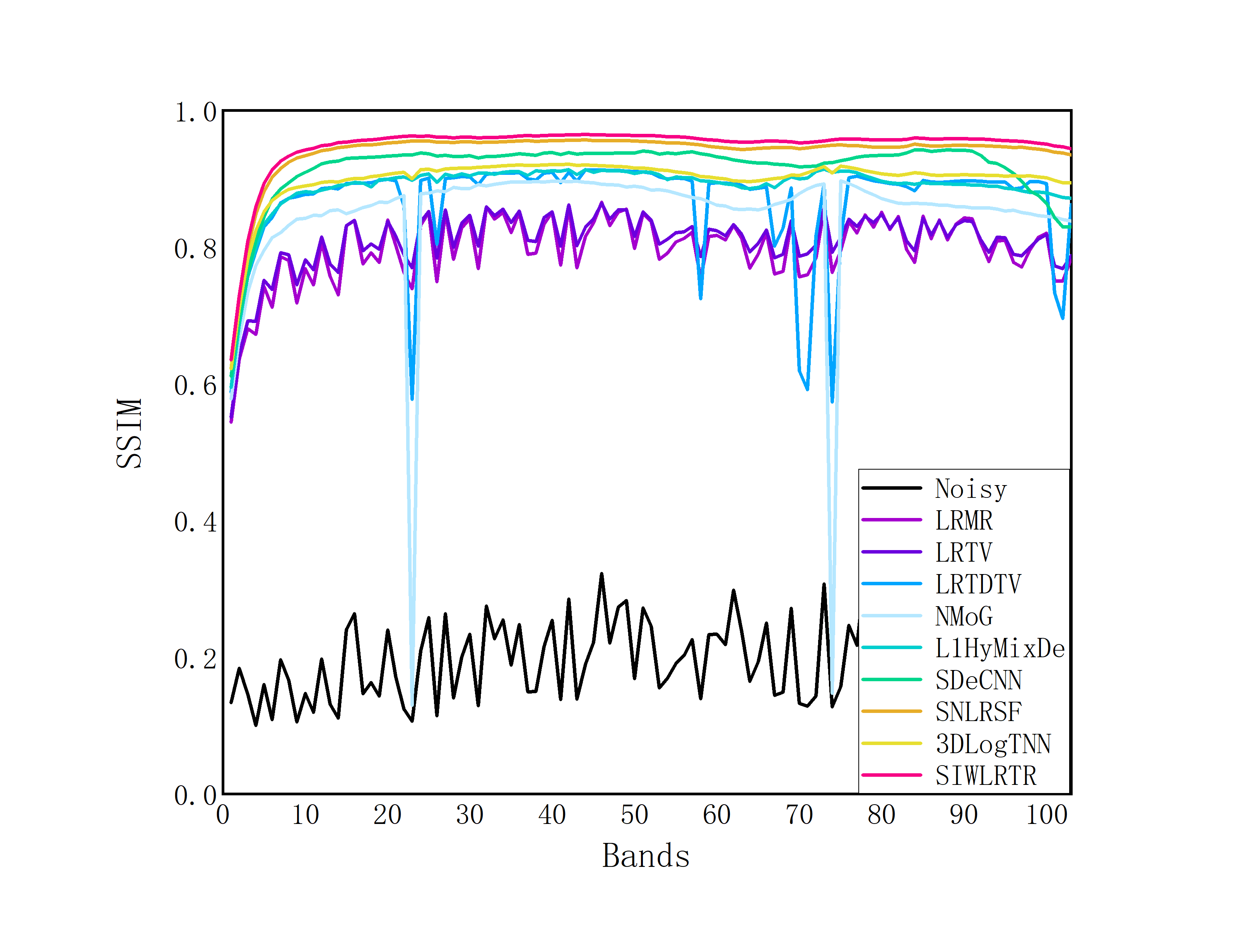}
    }
    \centering
    \caption{ PSNR and SSIM values of each band for PaviaU dataset in CASE1 and CASE2.(a) and (b) Case1.
              (c) and (d) Case2. }
    \label{Fig.4}
\end{figure*}

\begin{figure*}
    \centering
     \subfigure[]{
        \includegraphics[width=0.23\textwidth]{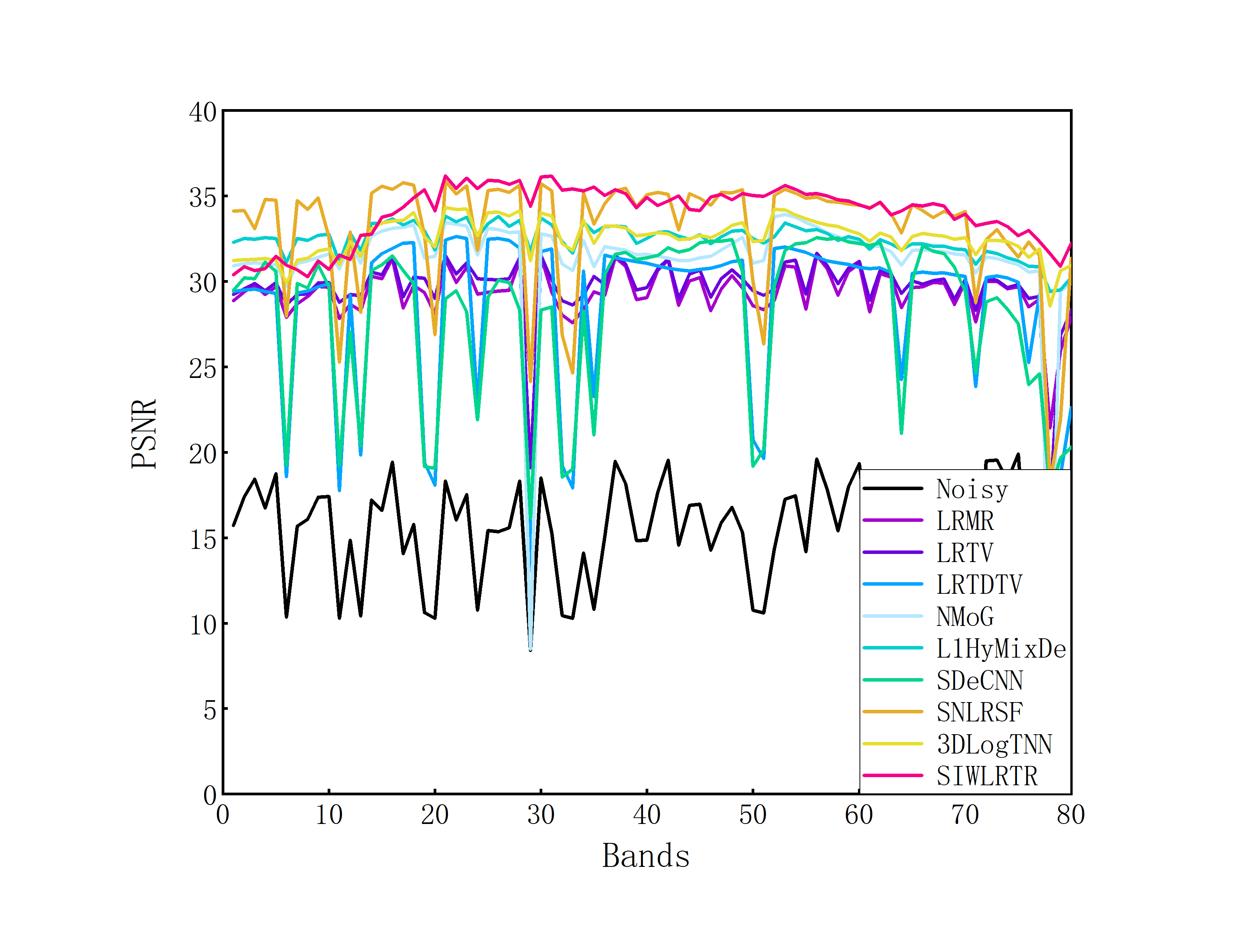}
    }
     \subfigure[]{
        \includegraphics[width=0.23\textwidth]{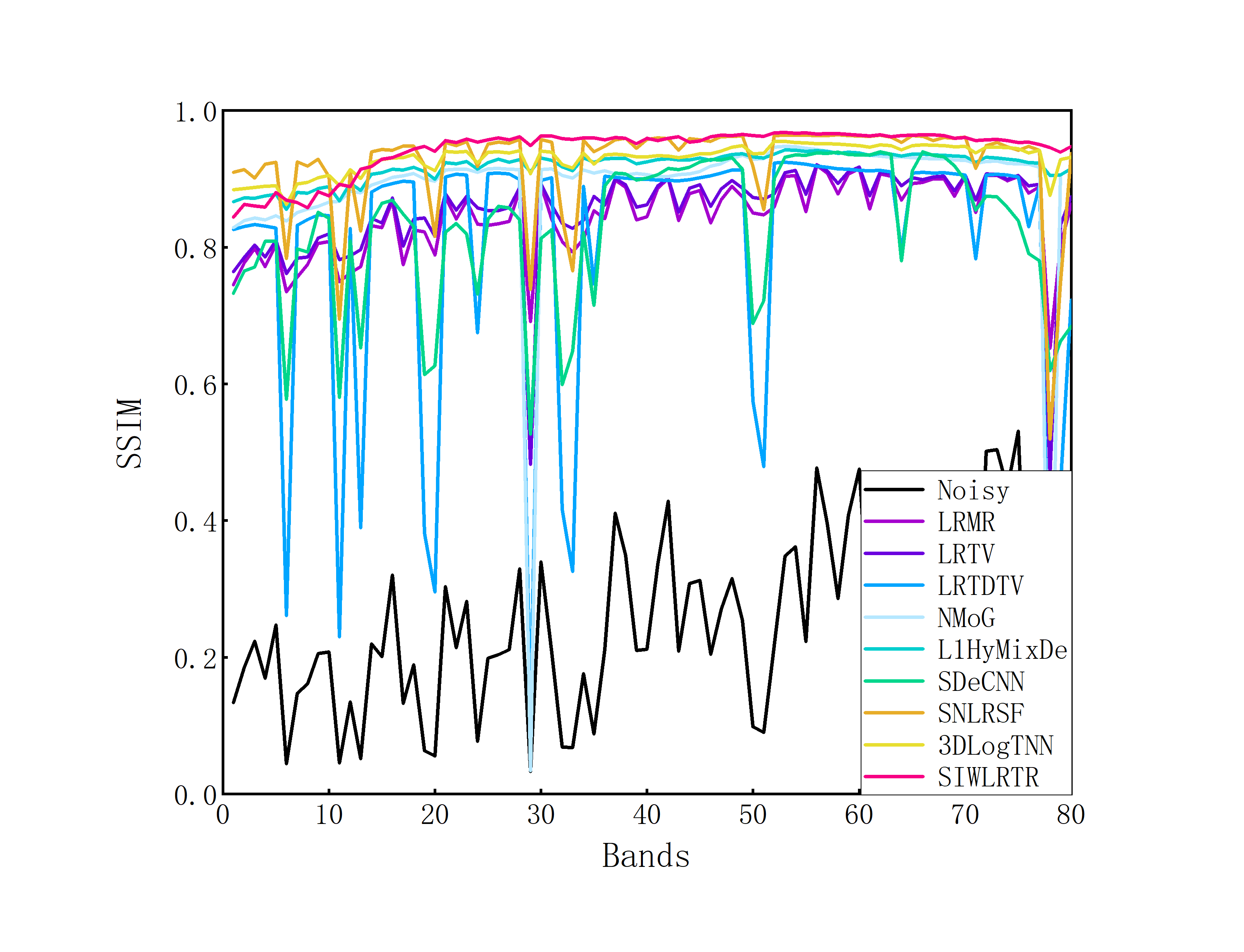}
    }
     \subfigure[]{
        \includegraphics[width=0.23\textwidth]{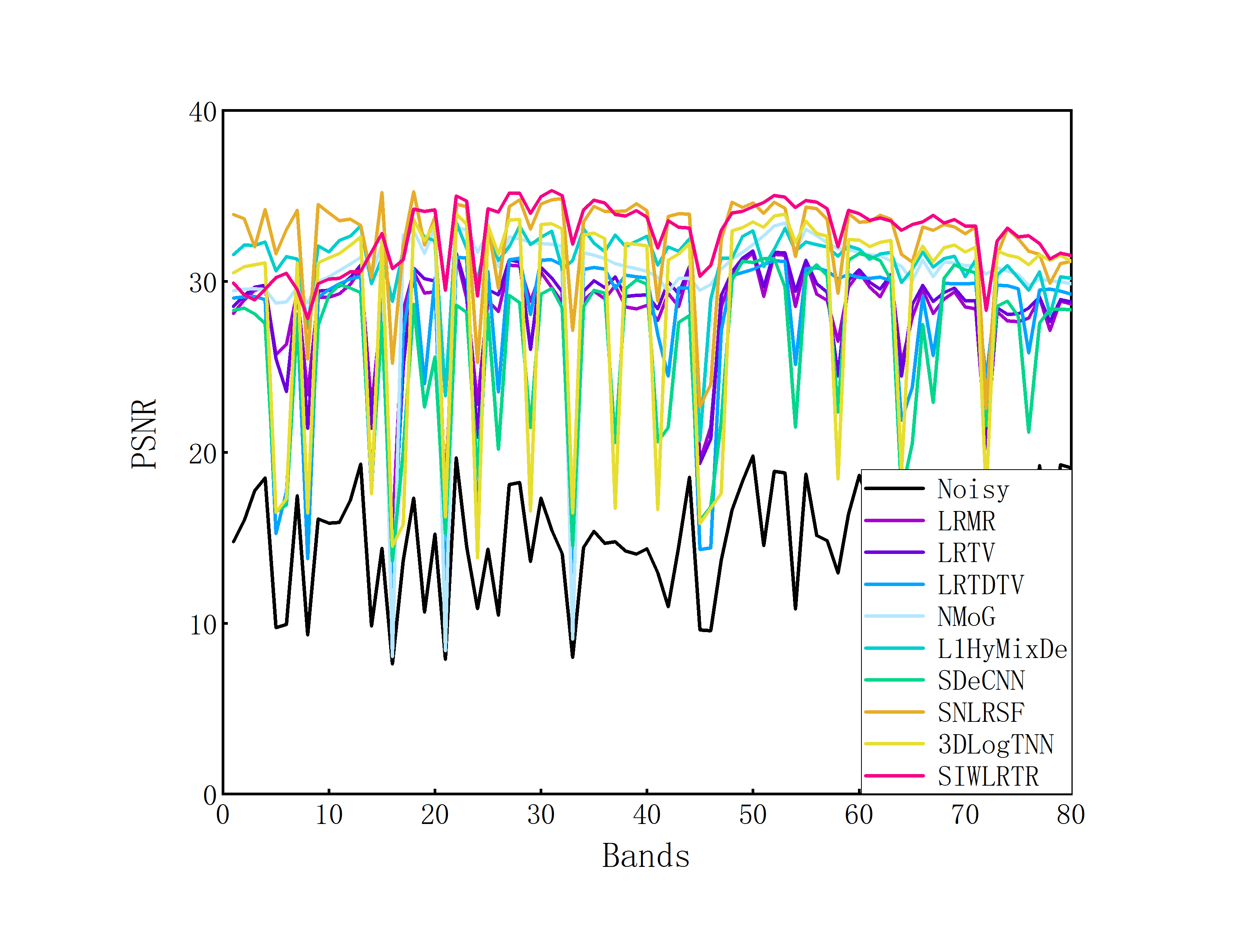}
    }
     \subfigure[]{
        \includegraphics[width=0.23\textwidth]{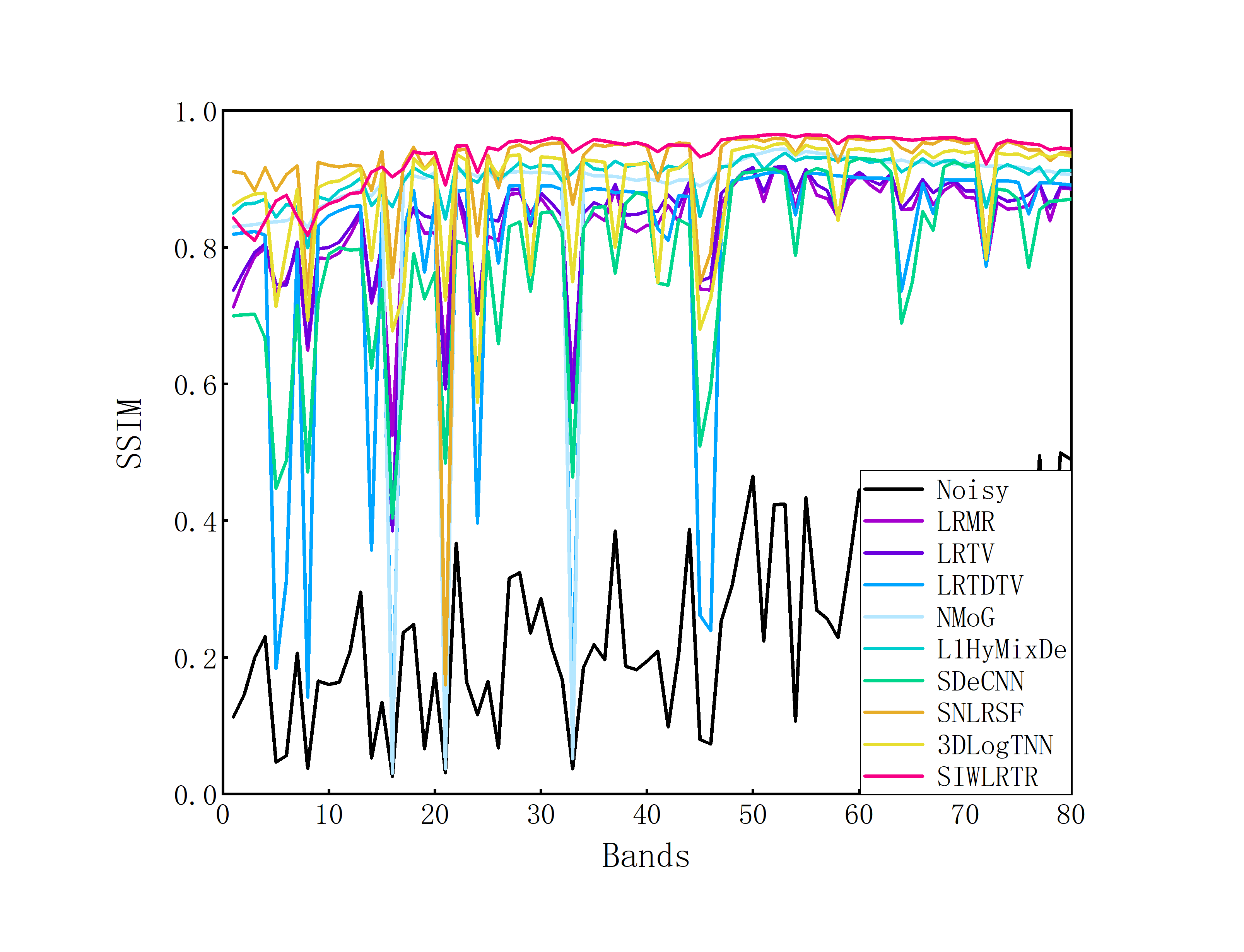}
    }
    \centering
    \caption{ PSNR and SSIM values of each band for PaviaC dataset in CASE3 and CASE4.(a) and (b) Case3.
              (c) and (d) Case4. }
    \label{Fig.5}
\end{figure*}

\begin{figure*}
    \centering
    \subfigure[Original band 60]{
        \includegraphics[width=0.13\textwidth]{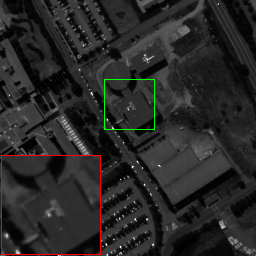}
    }
    \subfigure[Noisy]{
        \includegraphics[width=0.13\textwidth]{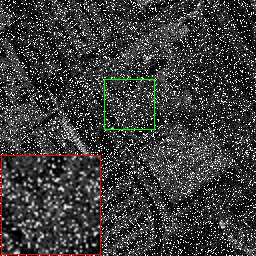}
    }
    \subfigure[LRMR]{
        \includegraphics[width=0.13\textwidth]{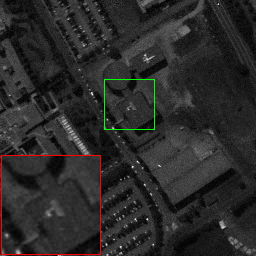}
    }
     \subfigure[LRTV]{
        \includegraphics[width=0.13\textwidth]{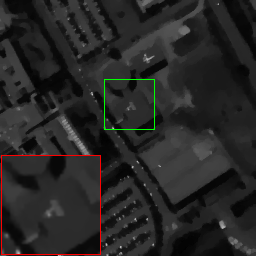}
    }
     \subfigure[LRTDTV]{
        \includegraphics[width=0.13\textwidth]{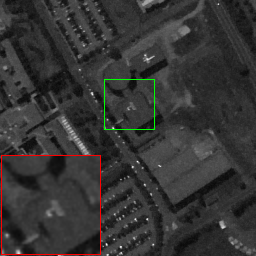}
    }
     \subfigure[NMoG]{
        \includegraphics[width=0.13\textwidth]{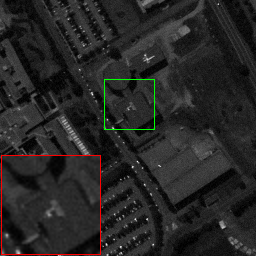}
    }
    \\
     \subfigure[L1HyMixDe]{
        \includegraphics[width=0.13\textwidth]{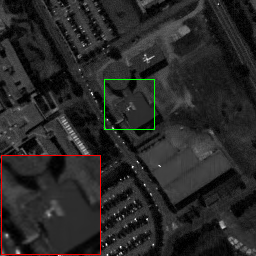}
    }
     \subfigure[SDeCNN]{
        \includegraphics[width=0.13\textwidth]{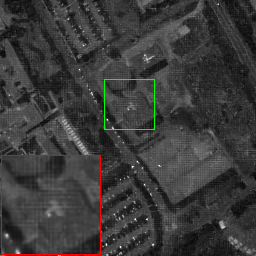}
    }
    \subfigure[SNLRSF]{
        \includegraphics[width=0.13\textwidth]{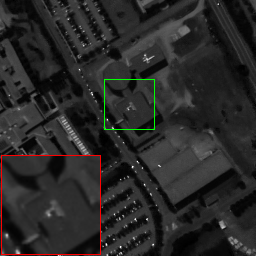}
    }
    \subfigure[3DLogTNN]{
        \includegraphics[width=0.13\textwidth]{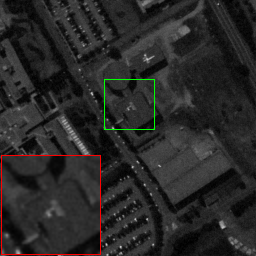}
    }
    \subfigure[SWLRTR]{
        \includegraphics[width=0.13\textwidth]{picture/U3result/U3_3DLogTNN.png}
    }
    \centering
    \caption{ Denoising results for PaviaU in CASE3. }
    \label{Fig.6}
\end{figure*}

\begin{figure*}
    \centering
    \subfigure[Original band 64]{
        \includegraphics[width=0.13\textwidth]{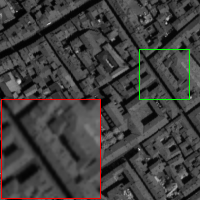}
    }
    \subfigure[Noisy]{
        \includegraphics[width=0.13\textwidth]{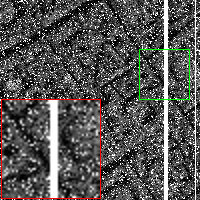}
    }
    \subfigure[LRMR]{
        \includegraphics[width=0.13\textwidth]{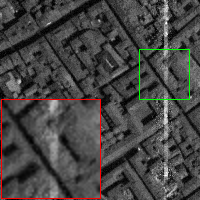}
    }
     \subfigure[LRTV]{
        \includegraphics[width=0.13\textwidth]{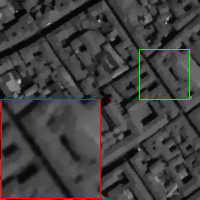}
    }
     \subfigure[LRTDTV]{
        \includegraphics[width=0.13\textwidth]{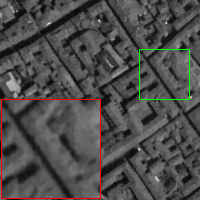}
    }
     \subfigure[NMoG]{
        \includegraphics[width=0.13\textwidth]{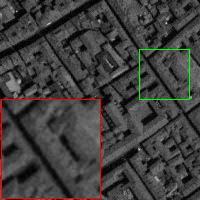}
    }
    \\
     \subfigure[L1HyMixDe]{
        \includegraphics[width=0.13\textwidth]{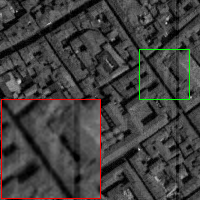}
    }
     \subfigure[SDeCNN]{
        \includegraphics[width=0.13\textwidth]{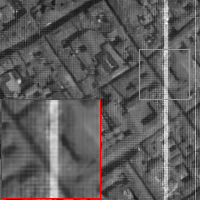}
    }
    \subfigure[SNLRSF]{
        \includegraphics[width=0.13\textwidth]{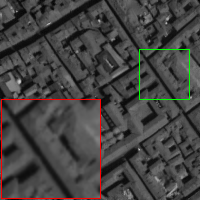}
    }
    \subfigure[3DLogTNN]{
        \includegraphics[width=0.13\textwidth]{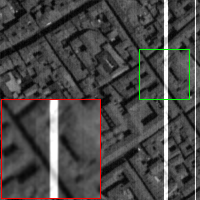}
    }
    \subfigure[SWLRTR]{
        \includegraphics[width=0.13\textwidth]{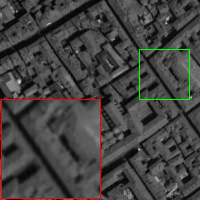}
    }
    \centering
    \caption{ Denoising results of PaviaC in CASE4. }
    \label{Fig.7}
\end{figure*}

\begin{figure}
    \centering
    \subfigure[]{
        \includegraphics[width=0.4\linewidth]{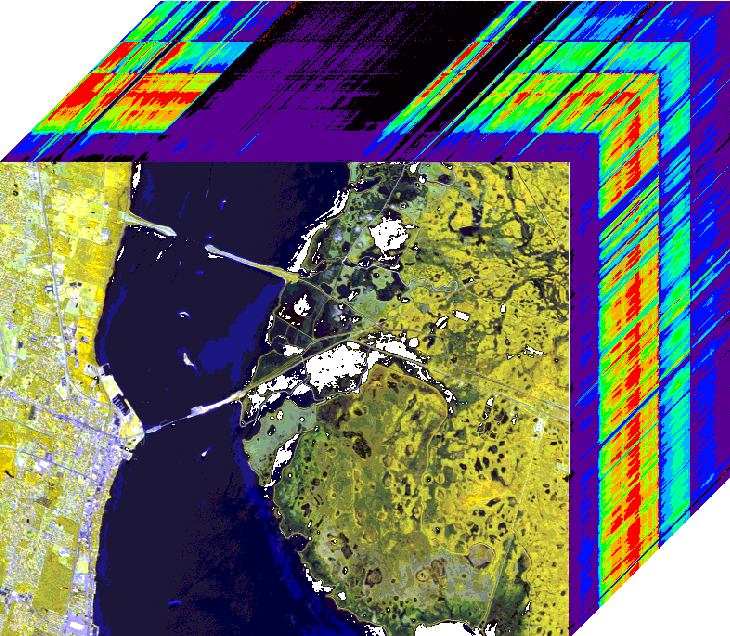}
    }
    \subfigure[]{
        \includegraphics[width=0.4\linewidth]{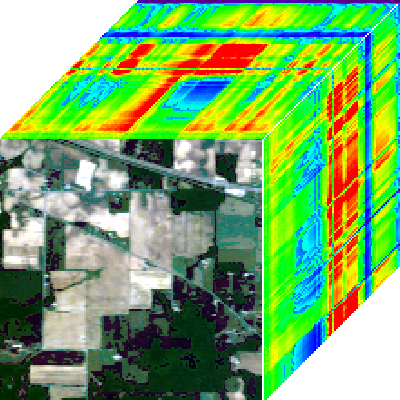}
    }
    \centering
    \caption{(a) Kennedy Space Center dataset (R:50,G:77,B:29) (b) Indian Pines (R:29,G:19,B:9)}
    \label{Fig.8}
\end{figure}

\begin{figure*}
    \centering
    \subfigure[Noisy band 89]{
        \includegraphics[width=0.18\textwidth]{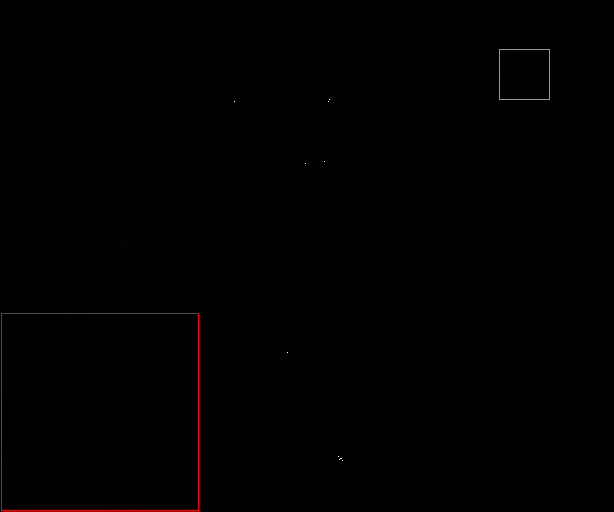}
    }
    \subfigure[LRMR]{
        \includegraphics[width=0.18\textwidth]{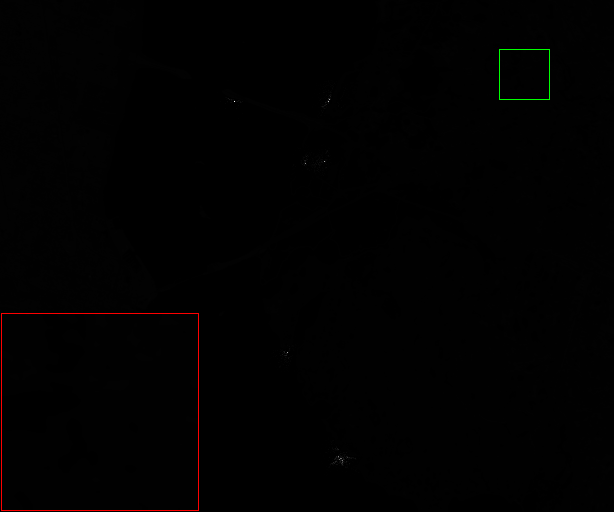}
    }
     \subfigure[LRTV]{
        \includegraphics[width=0.18\textwidth]{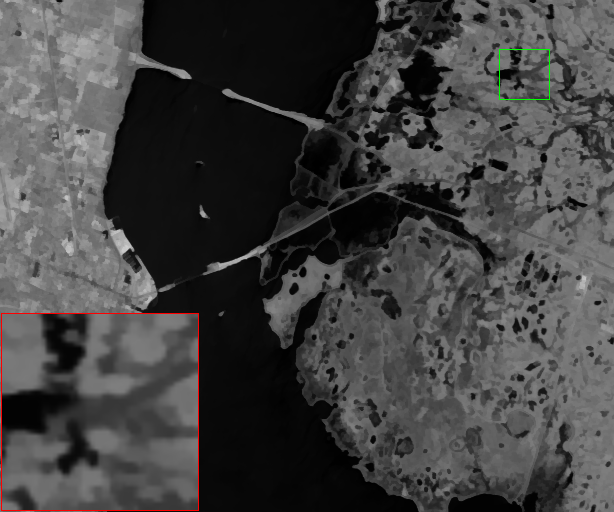}
    }
     \subfigure[LRTDTV]{
        \includegraphics[width=0.18\textwidth]{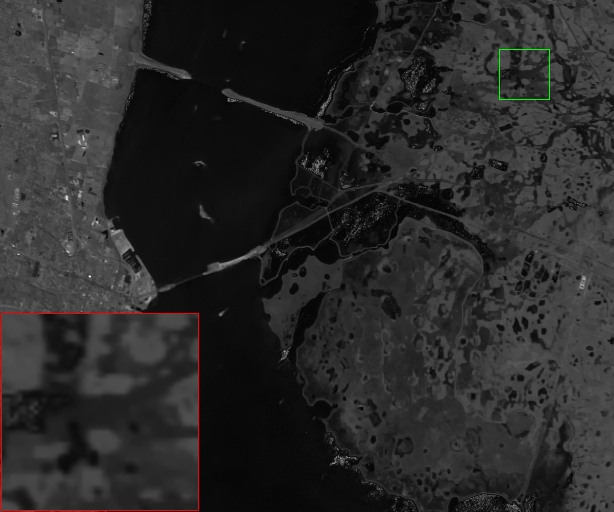}
    }
     \subfigure[NMoG]{
        \includegraphics[width=0.18\textwidth]{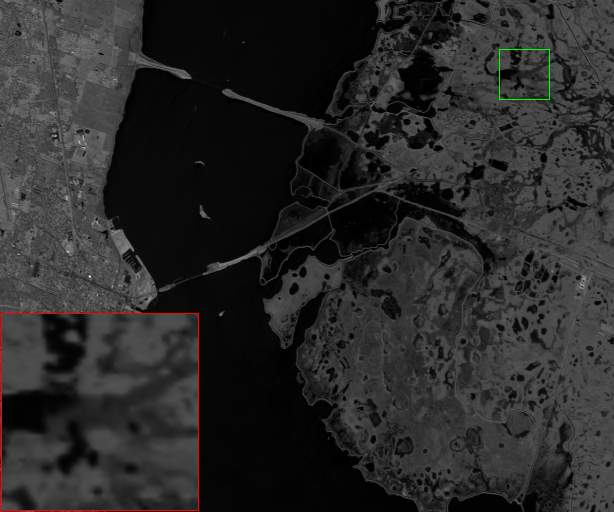}
    }
    \\
     \subfigure[L1HyMixDe]{
        \includegraphics[width=0.18\textwidth]{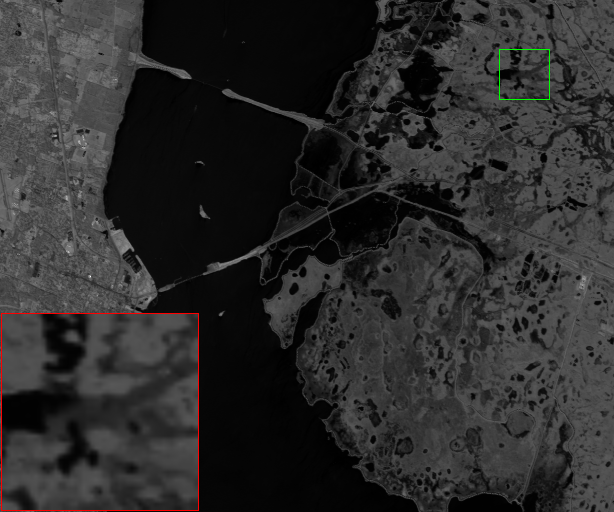}
    }
     \subfigure[SDeCNN]{
        \includegraphics[width=0.18\textwidth]{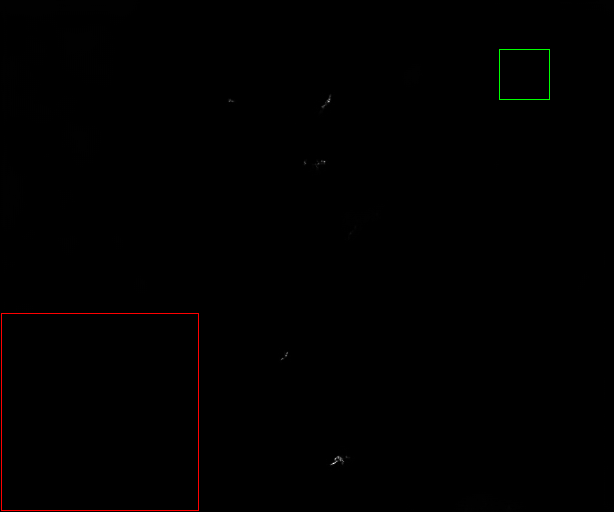}
    }
    \subfigure[SNLRSF]{
        \includegraphics[width=0.18\textwidth]{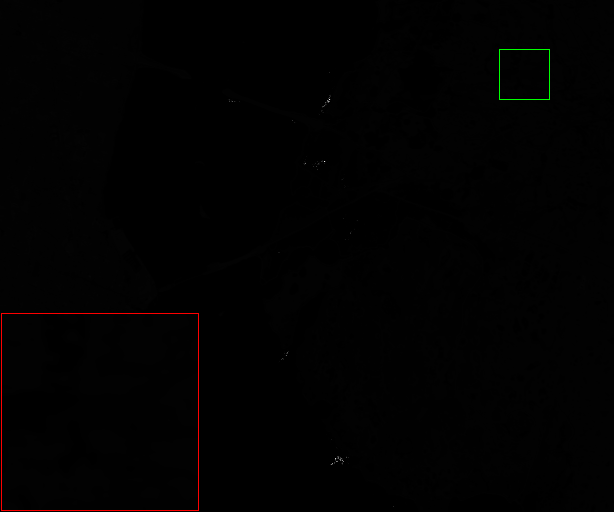}
    }
    \subfigure[3DLogTNN]{
        \includegraphics[width=0.18\textwidth]{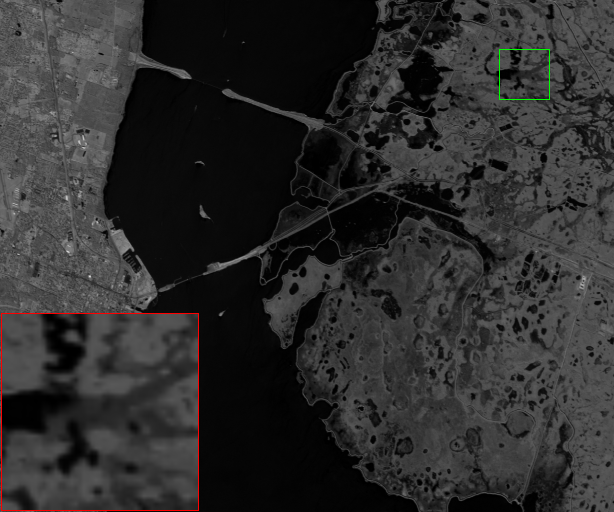}
    }
    \subfigure[SWLRTR]{
        \includegraphics[width=0.18\textwidth]{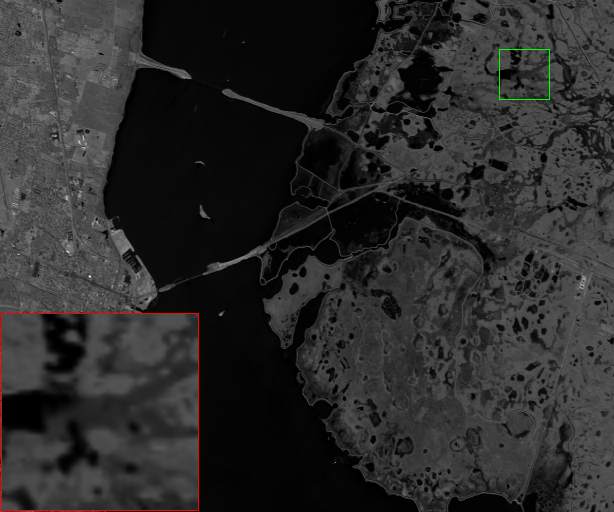}
    }
    \centering
    \caption{ Denoising results for KSC dataset. }
    \label{Fig.9}
\end{figure*}

\begin{figure*}
    \centering
    \subfigure[Noisy band 50]{
        \includegraphics[width=0.18\textwidth]{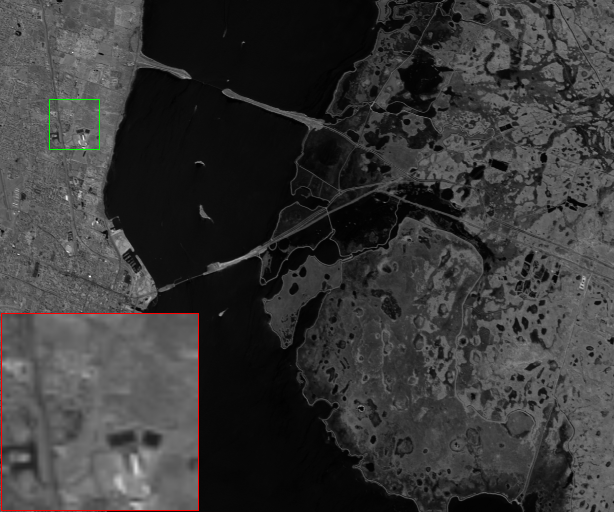}
    }
    \subfigure[LRMR]{
        \includegraphics[width=0.18\textwidth]{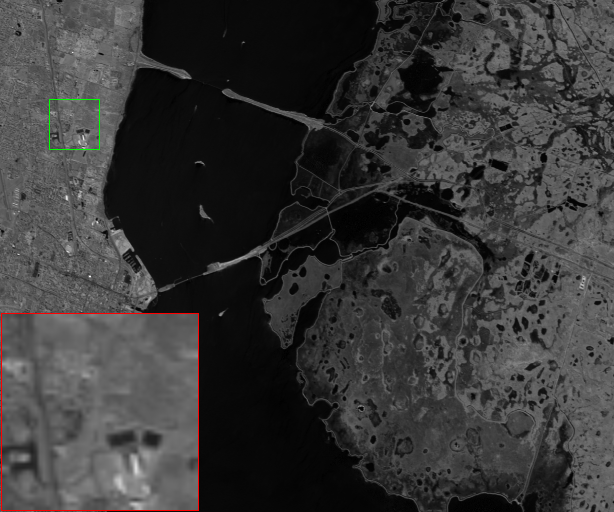}
    }
     \subfigure[LRTV]{
        \includegraphics[width=0.18\textwidth]{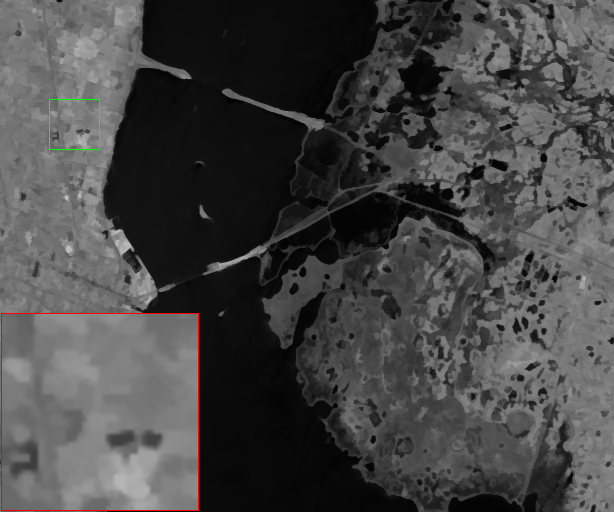}
    }
     \subfigure[LRTDTV]{
        \includegraphics[width=0.18\textwidth]{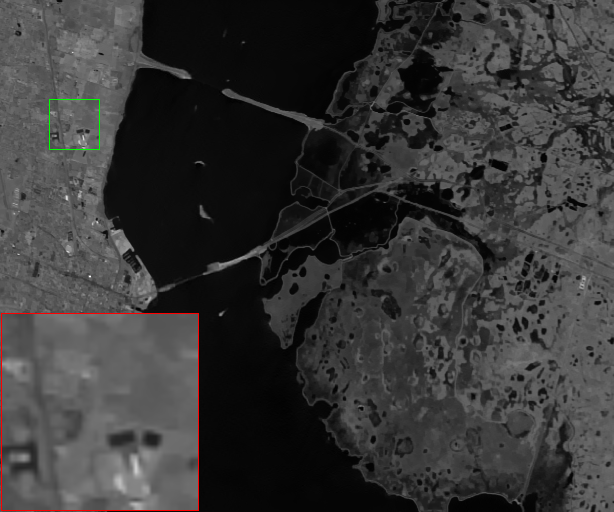}
    }
     \subfigure[NMoG]{
        \includegraphics[width=0.18\textwidth]{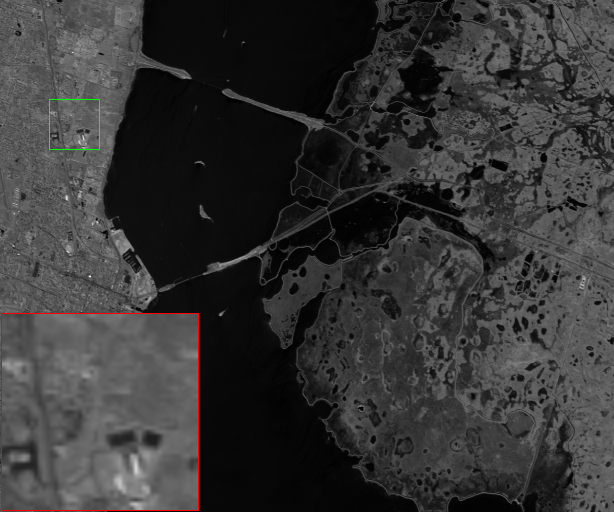}
    }
    \\
     \subfigure[L1HyMixDe]{
        \includegraphics[width=0.18\textwidth]{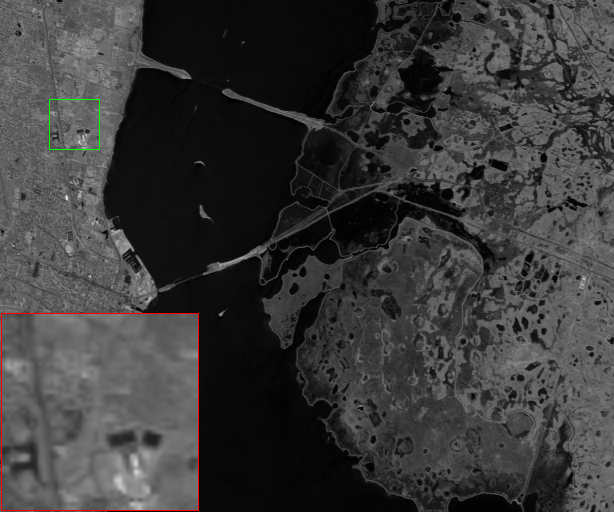}
    }
     \subfigure[SDeCNN]{
        \includegraphics[width=0.18\textwidth]{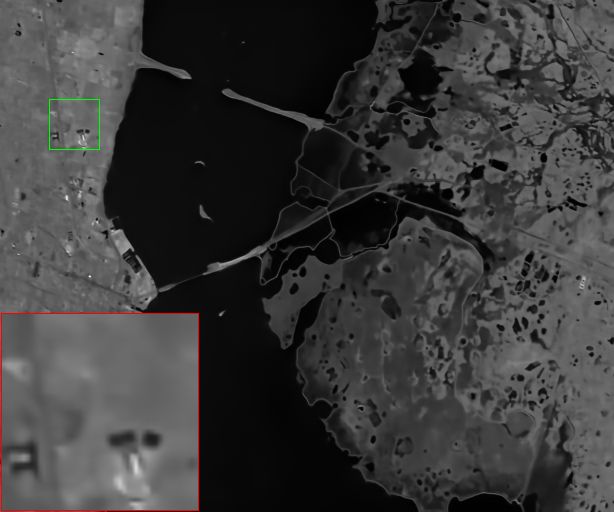}
    }
    \subfigure[SNLRSF]{
        \includegraphics[width=0.18\textwidth]{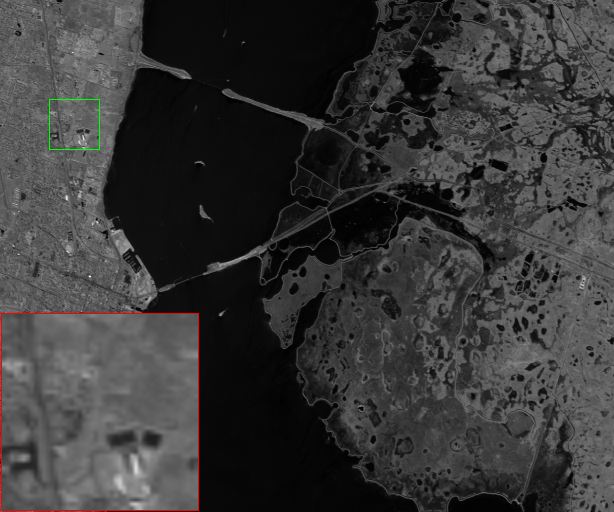}
    }
    \subfigure[3DLogTNN]{
        \includegraphics[width=0.18\textwidth]{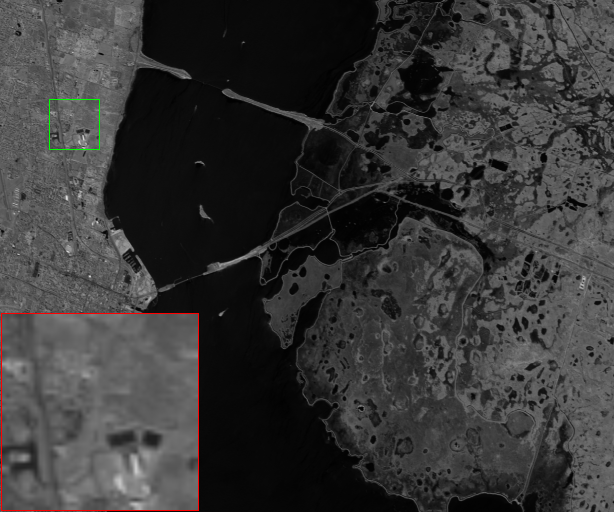}
    }
    \subfigure[SWLRTR]{
        \includegraphics[width=0.18\textwidth]{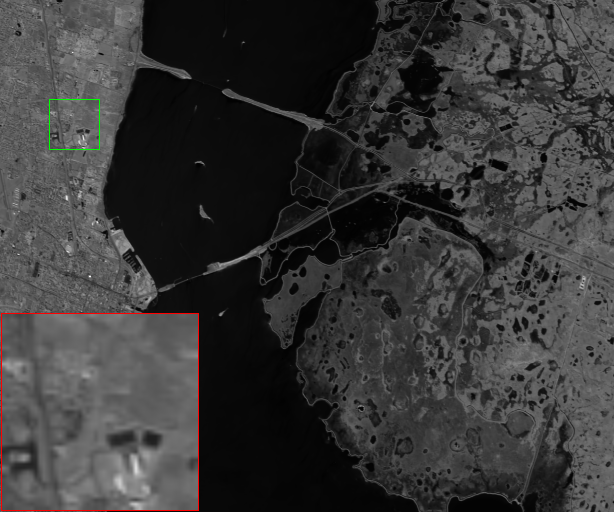}
    }
    \centering
    \caption{ Denoising results for KSC dataset. }
    \label{Fig.10}
\end{figure*}

\begin{figure*}
    \centering
    \subfigure[Noisy band 1]{
        \includegraphics[width=0.18\textwidth]{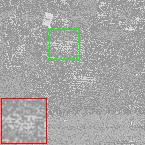}
    }
    \subfigure[LRMR]{
        \includegraphics[width=0.18\textwidth]{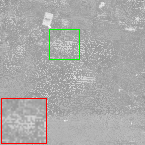}
    }
     \subfigure[LRTV]{
        \includegraphics[width=0.18\textwidth]{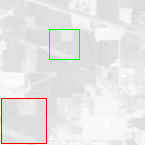}
    }
     \subfigure[LRTDTV]{
        \includegraphics[width=0.18\textwidth]{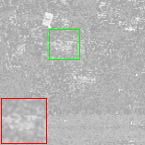}
    }
     \subfigure[NMoG]{
        \includegraphics[width=0.18\textwidth]{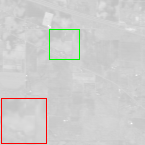}
    }
    \\
     \subfigure[L1HyMixDe]{
        \includegraphics[width=0.18\textwidth]{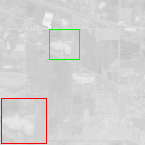}
    }
     \subfigure[SDeCNN]{
        \includegraphics[width=0.18\textwidth]{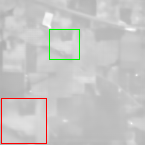}
    }
    \subfigure[SNLRSF]{
        \includegraphics[width=0.18\textwidth]{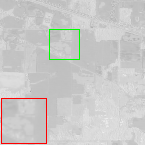}
    }
    \subfigure[3DLogTNN]{
        \includegraphics[width=0.18\textwidth]{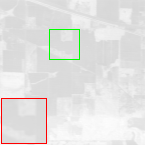}
    }
    \subfigure[SWLRTR]{
        \includegraphics[width=0.18\textwidth]{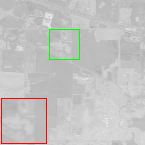}
    }
    \centering
    \caption{ Denoising results for Indian Pines dataset. }
    \label{Fig.11}
\end{figure*}

\begin{figure*}
    \centering
    \subfigure[]{
        \includegraphics[width=0.18\textwidth,height=2.5cm]{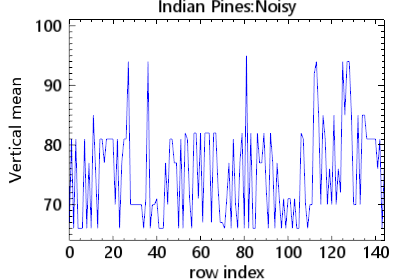}
    }
    \subfigure[]{
        \includegraphics[width=0.18\textwidth,height=2.5cm]{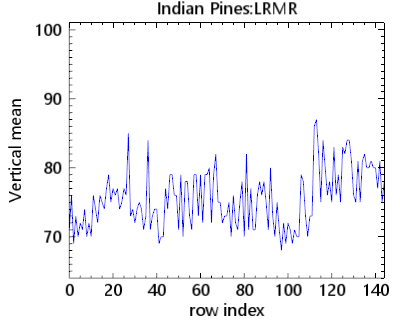}
    }
     \subfigure[]{
        \includegraphics[width=0.18\textwidth,height=2.5cm]{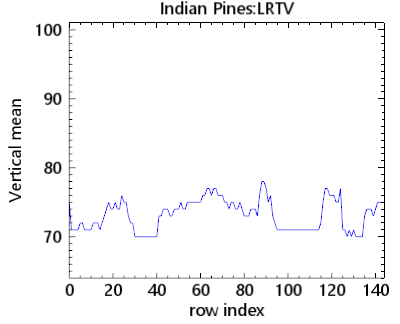}
    }
     \subfigure[]{
        \includegraphics[width=0.18\textwidth,height=2.5cm]{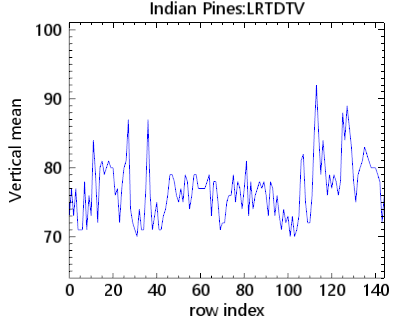}
    }
     \subfigure[]{
        \includegraphics[width=0.18\textwidth,height=2.5cm]{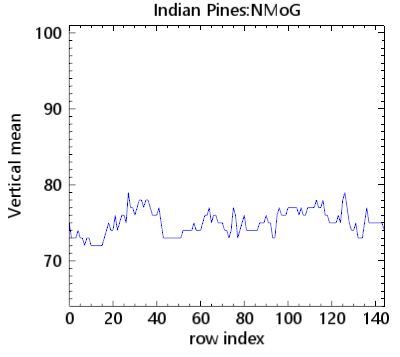}
    }
    \\
     \subfigure[]{
        \includegraphics[width=0.18\textwidth,height=2.5cm]{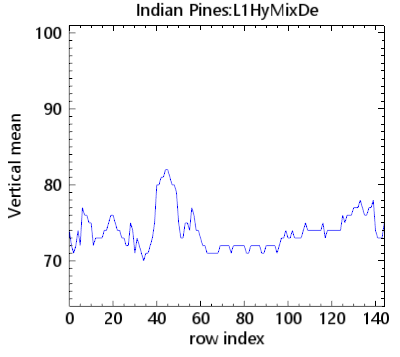}
    }
     \subfigure[]{
        \includegraphics[width=0.18\textwidth,height=2.5cm]{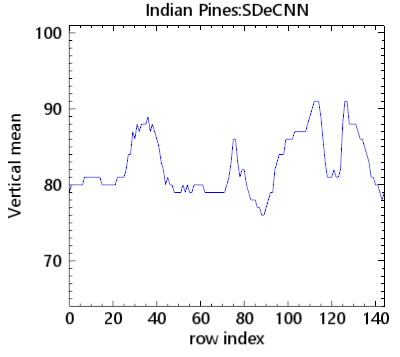}
    }
    \subfigure[]{
        \includegraphics[width=0.18\textwidth,height=2.5cm]{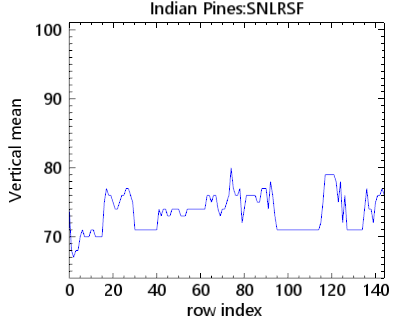}
    }
    \subfigure[]{
        \includegraphics[width=0.18\textwidth,height=2.5cm]{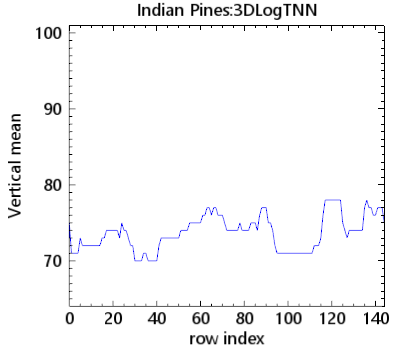}
    }
    \subfigure[]{
        \includegraphics[width=0.18\textwidth,height=2.5cm]{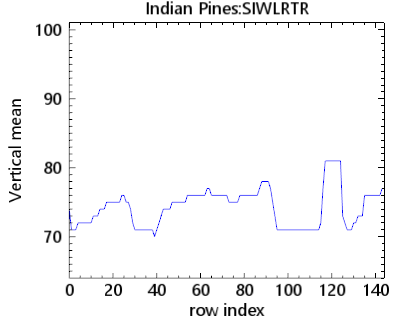}
    }
    \centering
    \caption{ Vertical mean profiles of band 28 in the real Indian Pines dataset experiment. }
    \label{Fig.12}
\end{figure*}

\subsection{ Iterative regularization }
Iterative regularization adds a specific proportion of the noise signal to the input of each iteration as a disturbance to the result. This strategy has been widely used to improve denoising performance. The formulation of updating the noisy image in the next iteration is
\begin{eqnarray}
{\mathcal{Y}_{n + 1}} = \alpha {\mathcal{X}_n} + (1 - \alpha )\mathcal{Y}
\end{eqnarray}
where $\alpha$ is a trade-off parameter that balances input in the next iteration, $n$ is the number of iterations. Subspace size $k$ is related to different datasets and noise intensity, we initialize $k$ according to HySime \cite{Bioucas2008Hyperspectral}. The noise intensity of each iteration becomes weaker as the denoising process goes on, the denoised image is closer to the real image. Fig.\ref{Fig.2} presents the PSNR values on different noise types and datasets with the increase of iteration, proves that $k$ should be gradually increased with the denoising process to retains more clean image information. Thus, $k$ is updated by
\begin{eqnarray}
k = k + \beta  \times n
\end{eqnarray}
where $\beta$ is a constant that determines the step size. In this way, ${A_{n + 1}}$ can extract more valuable information from the input image. The algorithm applied to solve the SWLRTR model is summarized as Algorithm 1, and the corresponding flowchart is shown as Fig.\ref{Fig.1}.

\subsection{ Complexity analysis }
In our algorithm, tucker decomposition and matrix multiplications operation dominate the computational complexity of SWLRTR. As for estimating ${\mathcal{Z}_i}$ of size ${p^2} \times k \times q$, it needs ${N_0} \times k \times \mathcal{O}(\min ({p^2}{k^2}{q^2},{p^4}kq))$ flops, where ${N_0}$ represents the number of iterations in low-rank tensor recovery. Hence, the process of obtain $\mathcal{Z}$ costs $np \times {N_0} \times k \times \mathcal{O}(\min ({p^2}{k^2}{q^2},{p^4}kq))$, where $np$ denotes the total number of patches search from HSI. After that, an amount of $k{n_1}{n_2}{n_3}$ and $\mathcal{O}(\min ({n_3}{k^2},n_3^2k))$ flops are required to calculate ${\mathcal{S}}$ and $A$ each update, respectively. The overall computational complexity of SWLRTR is $N \times (np \times {N_0} \times k \times \mathcal{O}(\min ({p^2}{k^2}{q^2},{p^4}kq)) + {N_1}(k{n_1}{n_2}{n_3} + \mathcal{O}(\min ({n_3}{k^2},n_3^2k))))$, where $N_1$ is the number of iterations needed for convergence when estimate ${\mathcal{S}}$, ${\mathcal{Z}}$ and $A$, $N$ defines total number of iterations for denoising.

\section{ experimental results }
We design simulated and real-data experiments to verify the strong capability of SWLRTR on removing mixed noise. In the experiment, we chose LRMR \cite{2014Hyperspectral}, LRTV \cite{2015Total}, LRTDTV \cite{0Hyperspectral}, NMoG \cite{2017Denoising}, L1HyMixDe \cite{2020Hyperspectral}, SDeCNN \cite{2020A}, 3DLogTNN \cite{2019Mixed}, SNLRSF \cite{DBLP:journals/staeors/CaoYZHXG19} as the comparison algorithms. All parameters are manually adjusted as recommended or according to the rules in the relevant literature to ensure optimal performance. For the convenience of calculation, we normalize the pixel values of HSI into [0,1]. The experiments are conducted in MATLAB R2018a using a 3.20-GHz CPU with 16-GB RAM. 

\subsection{ Simulated data experiments }
There are three HSI datasets adopted for simulated experiments, includes the Washington DC Mall, the Pavia University (PaviaU), and Pavia Center (PaviaC) datasets. The Washington DC Mall data contains $1208 \times 307$ pixels and 191 spectral channels, where a subimage of size $256 \times 256 \times 191$ is used in our experiments. The subimages of Pavia University (PaviaU) and Pavia Center (PaviaC) datasets are selected as $256 \times 256 \times 103$ and $200 \times 200 \times 80$, respectively. Fig.\ref{Fig.3} shows the three images.

Real HSIs are usually polluted by complex noise. To simulated the various noise distribution in the real scene, four types of noise are added to the original HSI data:
\\Case 1: Gaussian noise with the same distribution is added to each band of the three simulated datasets, whose mean is 0 and the noise variance $\sigma$ is set to 0.1.
\\Case 2: Gaussian noise with different intensity is added to each band of the three simulated datasets, whose mean is 0 and the range of noise variance $\sigma$ is [0.1,0.2]. 
\\Case 3: Gaussian noise the same as Case 2 is added to each band of the three simulated datasets. Besides, 20 bands are chosen randomly to add impulse noise with a percentage of 20\%.
\\Case 4: Gaussian noise and impulse noise the same as Case 3 are added to the original image. We select 10 bands from the impulse noise bands and the other 10 bands from the rest to add deadlines width of 1-3.

\begin{table*}
\caption{Quantitative evaluation of different methods on the simulated data in different cases}
\centering
\resizebox{0.95\textwidth}{!}
{\begin{tabular}{@{}ccccccccccccc@{}}
\toprule
Data                     & Noise case              & Metrics & Noisy   & LRMR    & LRTV    & LRTDTV  & NMoG    & L1HyMixDe & SDeCNN  & SNLRSF          & 3DLogTNN & SWLRTR          \\ \midrule
                         &                         & MPSNR   & 20.0003 & 34.2396 & 31.0162 & 35.1201 & 35.8342 & 35.8338   & 35.4117 & 38.3837         & 35.8497  & \textbf{38.5452} \cr
                         & {Case 1}                & MSSIM   & 0.4218  & 0.9384  & 0.8689  & 0.9551  & 0.9563  & 0.9525    & 0.9547  & 0.9774          & 0.9641   & \textbf{0.9791}  \cr 
                         &                         & ERGAS   & 388.501 & 72.84   & 106.019 & 66.601  & 61.229  & 60.546    & 64.993  & 45.8951         & 60.519   & \textbf{45.019}  \cr 
                         &                         & MSA     & 0.461   & 0.092   & 0.086   & 0.065   & 0.071   & 0.069     & 0.077   & 0.0503          & 0.062    & \textbf{0.049}   \cr 
\cline{2-13}             
                         &                         & MPSNR   & 16.6878 & 31.5429 & 29.2826 & 34.0946 & 33.7798 & 33.6042   & 33.1842 & 36.7837         & 34.1315  & \textbf{36.9951} \cr 
                         & {Case 2}                & MSSIM   & 0.2851  & 0.895   & 0.8093  & 0.9404  & 0.9315  & 0.9214    & 0.9274  & 0.9671          & 0.9466   & \textbf{0.97}    \cr
                         &                         & ERGAS   & 586.274 & 99.492  & 129.256 & 74.092  & 76.933  & 77.918    & 83.312  & 54.4514         & 73.262   & \textbf{53.131}  \cr
{WDC}                         &                         & MSA     & 0.633   & 0.127   & 0.104   & 0.077   & 0.09    & 0.088     & 0.096   & 0.0581          & 0.074    & \textbf{0.055}   \cr
\cline{2-13}                         
                         &                         & MPSNR   & 16.0763 & 31.2493 & 29.1227 & 32.1896 & 33.4951 & 33.145    & 30.9073 & 36.2763         & 33.9541  & \textbf{36.5271} \cr 
                         &  {Case 3}               & MSSIM   & 0.2696  & 0.8897  & 0.8033  & 0.9177  & 0.9252  & 0.9145    & 0.8848  & 0.9636          & 0.9444   & \textbf{0.9669}  \cr
                         &                         & ERGAS   & 683.055 & 102.958 & 131.649 & 111.133 & 79.706  & 82.643    & 136.536 & 57.4614         & 74.84    & \textbf{55.805}  \cr
                         &                         & MSA     & 0.673   & 0.132   & 0.105   & 0.127   & 0.094   & 0.095     & 0.163   & 0.0603          & 0.075    & \textbf{0.057}   \cr 
\cline{2-13}                         
                         &                         & MPSNR   & 15.9527 & 31.0065 & 29.154  & 31.2589 & 32.5444 & 33.0966   & 29.9408 & 36.2179         & 32.1136  & \textbf{36.3291} \cr
                         & {Case 4}                & MSSIM   & 0.2672  & 0.8857  & 0.8055  & 0.9052  & 0.9136  & 0.9134    & 0.8612  & 0.9633          & 0.9256   & \textbf{0.966}   \cr
                         &                         & ERGAS   & 711.236 & 110.402 & 131.051 & 163.731 & 229.482 & 85.213    & 194.206 & 58.0860         & 178.079  & \textbf{58.158}  \cr 
                         &                         & MSA     & 0.682   & 0.14    & 0.105   & 0.174   & 0.211   & 0.1       & 0.214   & 0.0617          & 0.176    & \textbf{0.061}   \cr
\cline{1-13}                        
                         &                         & MPSNR   & 20.0002 & 32.8386 & 32.3885 & 33.5039 & 34.4763 & 34.8787   & 36.4965 & 37.4480         & 34.6318  & \textbf{37.8838} \cr
                         & {Case 1}                 & MSSIM   & 0.3226  & 0.8728  & 0.8841  & 0.9031  & 0.9061  & 0.9178    & 0.9431  & 0.9546          & 0.9301   & \textbf{0.9595}  \cr
                         &                         & ERGAS   & 396.159 & 91.374  & 94.852  & 83.616  & 77.075  & 72.405    & 60.934  & 55.4246         & 74.662   & \textbf{52.634}  \cr
                         &                         & MSA     & 0.527   & 0.129   & 0.091   & 0.093   & 0.1     & 0.089     & 0.079   & 0.0684          & 0.079    & \textbf{0.065}   \cr
\cline{2-13}                        
                         &                         & MPSNR   & 16.8769 & 30.3415 & 30.7691 & 32.13   & 32.5218 & 33.0366   & 34.4312 & 35.9842         & 33.2557  & \textbf{36.5373} \cr 
                         & {Case 2}                & MSSIM   & 0.2124  & 0.7983  & 0.8449  & 0.8663  & 0.8626  & 0.8834    & 0.9163  & 0.9397          & 0.9001   & \textbf{0.9484}  \cr
                         &                         & ERGAS   & 589.422 & 121.27  & 113.664 & 101.972 & 94.846  & 88.362    & 76.217  & 64.2431         & 86.487   & \textbf{60.25}   \cr
{PaviaU}                 &                         & MSA     & 0.697   & 0.168   & 0.101   & 0.127   & 0.118   & 0.105     & 0.096   & 0.0759          & 0.091    & \textbf{0.07}    \cr
\cline{2-13}                        
                         &                         & MPSNR   & 15.3987 & 29.6441 & 29.9775 & 29.2625 & 31.7338 & 32.0287   & 29.3647 & 34.4958         & 32.6137  & \textbf{35.2146} \cr 
                         & {Case 3}                & MSSIM   & 0.1742  & 0.7757  & 0.828   & 0.7866  & 0.8438  & 0.8562    & 0.8137  & 0.9243          & 0.8825   & \textbf{0.937}   \cr 
                         &                         & ERGAS   & 746.995 & 130.955 & 124.891 & 194.806 & 102.836 & 98.805    & 195.675 & 75.8599         & 93.171   & \textbf{70.273}  \cr 
                         &                         & MSA     & 0.768   & 0.172   & 0.11    & 0.238   & 0.125   & 0.115     & 0.234   & 0.085           & 0.102    & \textbf{0.08}    \cr
\cline{2-13}                         
                         &                         & MPSNR   & 15.2812 & 29.2353 & 29.1979 & 28.192  & 31.6299 & 32.1282   & 28.6306 & 33.8297         & 29.8944  & \textbf{34.3585} \cr
                         & {Case 4}                & MSSIM   & 0.1801  & 0.7742  & 0.8113  & 0.7674  & 0.8456  & 0.8617    & 0.7837  & 0.916           & 0.8454   & \textbf{0.9282}  \cr 
                         &                         & ERGAS   & 790.317 & 151.634 & 160.176 & 272.168 & 115.189 & 100.156   & 266.319 & 89.4595         & 257.499  & \textbf{82.77}   \cr 
                         &                         & MSA     & 0.774   & 0.192   & 0.176   & 0.301   & 0.15    & 0.119     & 0.295   & 0.1042          & 0.268    & \textbf{0.098}   \cr
\cline{1-13}                        
                         &                         & MPSNR   & 19.9985 & 32.5801 & 33.1567 & 34.0404 & 34.6891 & 34.9898   & 36.0434 & 37.4129         & 34.8657  & \textbf{37.6764} \cr
                         & {Case 1}                & MSSIM   & 0.436   & 0.9197  & 0.9239  & 0.94    & 0.9482  & 0.951     & 0.9618  & 0.9726          & 0.9579   & \textbf{0.9744}  \cr 
                         &                         & ERGAS   & 368.977 & 87.532  & 80.318  & 72.838  & 69.398  & 65.703    & 58.248  & 49.6967         & 68.15    & \textbf{48.199}  \cr
                         &                         & MSA     & 0.549   & 0.164   & 0.081   & 0.115   & 0.113   & 0.103     & 0.088   & \textbf{0.0688} & 0.082    & 0.069            \cr 
\cline{2-13}                         
                         &                         & MPSNR   & 16.8626 & 29.9801 & 31.311  & 32.251  & 32.4967 & 32.8791   & 33.7069 & 35.6506         & 33.1611  & \textbf{35.9845} \cr 
                         & {Case 2}                & MSSIM   & 0.2959  & 0.864   & 0.8882  & 0.8997  & 0.9207  & 0.9243    & 0.9366  & 0.9598          & 0.9389   & \textbf{0.9631}  \cr
                         &                         & ERGAS   & 553.476 & 119.532 & 99.474  & 103.478 & 88.999  & 83.816    & 76.676  & 60.8950         & 82.633   & \textbf{58.738}  \cr 
{PaviaC}                         &                         & MSA     & 0.705   & 0.216   & 0.092   & 0.178   & 0.138   & 0.125     & 0.108   & 0.0787          & 0.095    & \textbf{0.074}   \cr 
\cline{2-13}                        
                         &                         & MPSNR   & 15.4409 & 29.3152 & 29.8296 & 28.4451 & 31.6063 & 32.4358   & 28.1541 & 33.0666         & 32.5788  & \textbf{33.9795} \cr 
                         & {Case 3}                & MSSIM   & 0.2477  & 0.8456  & 0.8564  & 0.8011  & 0.9066  & 0.9172    & 0.8297  & 0.9236          & 0.9304   & \textbf{0.9432}  \cr 
                         &                         & ERGAS   & 716.014 & 130.482 & 167.091 & 225.046 & 106.95  & 88.382    & 209.581 & 105.1341        & 88.584   & \textbf{78.871}  \cr 
                         &                         & MSA     & 0.757   & 0.199   & 0.208   & 0.293   & 0.148   & 0.121     & 0.251   & 0.1579          & 0.103    & \textbf{0.109}   \cr
\cline{2-13}                         
                         &                         & MPSNR   & 14.7497 & 28.0716 & 29.711  & 27.3758 & 30.5504 & 31.5816   & 26.1214 & 32.2624         & 28.7468  & \textbf{32.7317} \cr
                         & {Case 4}                & MSSIM   & 0.2344  & 0.8298  & 0.853   & 0.7872  & 0.8882  & 0.9085    & 0.7854  & 0.9158          & 0.8859   & \textbf{0.9329}  \cr 
                         &                         & ERGAS   & 789.787 & 191.133 & 190.804 & 315.467 & 218.072 & 103.195   & 282.831 & 121.3199        & 286.35   & \textbf{92.804}  \cr
                         &                         & MSA     & 0.784   & 0.256   & 0.205   & 0.356   & 0.25    & 0.131     & 0.315   & 0.1682          & 0.304    & \textbf{0.121}   \cr
\cline{1-13}
\end{tabular}}
\end{table*}

\begin{table*}
\centering
\caption{Classification results on Kennedy Space Center(KSC) dataset with different denoising methods}
\begin{tabular}{@{}ccccccccccccc@{}}
\hline
Class           & Noisy  & LRMR   & LRTV   & LRTDTV & NMoG   & L1HyMixDe & SDeCNN & SNLRSF & 3DLogTNN & SWLRTR \\ \hline
Scrub           & 0.8333 & 0.8333 & 0.9868 & 0.9481 & 1      & 0.9481    & 1      & 1      & 1        & 1       \\ \hline
Willow swamp    & 0.7667 & 0.7667 & 1      & 0.8571 & 0.8846 & 1         & 0.8889 & 1      & 1        & 1       \\ \hline
CP hammock      & 0.697  & 0.6098 & 0.9615 & 0.7826 & 0.8621 & 1         & 1      & 0.8571 & 0.96     & 0.9615  \\ \hline
CP/Oak          & 0.3929 & 1      & 0.7667 & 0.5313 & 0.697  & 0.84      & 1      & 0.84   & 0.6765   & 1       \\ \hline
Slash pine      & 0.8000 & 1      & 0.8333 & 0.6667 & 0.8182 & 0.6667    & 0.8421 & 0.8571 & 0.75     & 1       \\ \hline
Oak/Broadleaf   & 0.6667 & 0.6923 & 0.9524 & 0.5789 & 0.85   & 0.7826    & 1      & 0.8462 & 0.9167   & 0.9200  \\ \hline
Hardwood swamp  & 1.0000 & 0.9167 & 1      & 0.9091 & 0.9091 & 0.9167    & 1.0000 & 1      & 1        & 1       \\ \hline
Graminoid marsh & 0.7778 & 0.8205 & 1      & 0.7857 & 0.9286 & 0.9762    & 0.9545 & 0.9091 & 0.9762   & 1       \\ \hline
Spartina marsh  & 0.8276 & 0.8276 & 0.963  & 0.8070 & 0.963  & 0.9455    & 0.9808 & 0.9808 & 0.9455   & 0.9811  \\ \hline
Catiail marsh   & 1      & 1      & 1      & 0.8043 & 1      & 1         & 1      & 1      & 1        & 1       \\ \hline
Salt marsh      & 1      & 1      & 1      & 0.9737 & 1      & 1         & 1      & 1      & 1        & 1       \\ \hline
Mud flats       & 0.8103 & 0.8909 & 1      & 0.8333 & 0.9796 & 1         & 1      & 1      & 1        & 1       \\ \hline
Water           & 1      & 1      & 1      & 0.9783 & 1      & 1         & 1      & 1      & 1        & 1       \\ \hline
OA              & 0.8429 & 0.8755 & 0.9732 & 0.8467 & 0.9444 & 0.9559    & 0.9828 & 0.9636 & 0.9617   & \textbf{0.9923}  \\ \hline
AA              & 0.8132 & 0.8737 & 0.9587 & 0.8043 & 0.9148 & 0.9289    & 0.9743 & 0.9454 & 0.9404   & \textbf{0.9894}  \\ \hline
Kappa           & 0.8247 & 0.8611 & 0.9701 & 0.8293 & 0.9382 & 0.9509    & 0.9808 & 0.9595 & 0.9574   & \textbf{0.9915}  \\ \hline
\end{tabular}
\end{table*}

\begin{table*}
\centering
\caption{ running time of different denoising methods on the three simulated data }
\begin{tabular}{@{}ccccccccccccc@{}}
\hline
Dataset & LRMR & LRTV        & LRTDTV & NMoG & L1HyMixDe & SDeCNN      & SNLRSF & 3DLogTNN & SWLRTR      \\ \hline
WDC     & 48   & \textbf{46} & 154    & 144  & 234       & 60          & 400    & 442      & 70          \\ \hline
PaviaU  & 34   & \textbf{33} & 85     & 104  & 244       & \textbf{33} & 407    & 189      & 69          \\ \hline
PaviaC  & 19   & 18          & 43     & 53   & 147       & \textbf{14} & 243    & 77       & \textbf{14} \\ \hline
\end{tabular}
\end{table*}

\subsubsection{ Quantitative comparison }
In this paper, four quantitative evaluation indexes are used to measure denoising performance: Mean peak signal-to-noise ratio (MPSNR), which is the average value of PSNR over all bands. PSNR evaluates the similarity between the clean image and denoising image by calculating the mean square error (MSE). Mean structural similarity (MSSIM), which is the average value of SSIM over all bands. SSIM is an evaluation index that measures the structural consistency of images before and after denoising. The erreur relative globale adimensionnelle de synthèse (ERGAS) is a global statistical measure based on the weighted sum of the mean square errors of each band. In addition, mean spectral angle mapping (MSAM) calculates the average of the spectral angles, which represents the similarity between spectra. Among them, higher MPSNR and MSSIM and lower ERGAS and MSA suggest better denoising results.

Table I shows the experimental results of the proposed algorithm and the comparison algorithms on the three datasets, including the values of different evaluation indicators in the four simulated noise situations Case 1-4. The best results are in bold. The denoising results of SWLRTR listed in the last column are bolded except the MSA measure of PaviaU in Case 4, which indicates the proposed method combining weighted low-rank tensor and subspace representation achieves competitive performance compared with other algorithms under the four quantitative index. In order to further demonstrate the denoising effect in each band, Fig.\ref{Fig.4} and \ref{Fig.5} display the PSNR and SSIM value in all bands under four kinds of simulated noises on Pavia University and Pavia Center datasets, where the red line represents the SWLRTR. It can be seen that the red line is higher than others in most bands and cases. Thus, the superiority of this method is verified quantitatively.

\subsubsection{ Visual comparison }
Fig.\ref{Fig.6} and Fig.\ref{Fig.7} show the original image, analog noise image, and the denoising results of different methods to compare the denoising performance of different methods visually. Specifically, on the PaviaU dataset in Fig.\ref{Fig.6}, band 60 is selected as a representative band because it includes Gaussian noise and impulse noise. Fig.\ref{Fig.7} show the recovery results in band 64 destroyed by Gaussian noise, impulse noise, and deadlines on PaviaC dataset. It can be seen that the SWLRTR retains the details and removes the unexpected mixed noise well compared with other algorithms in Fig.\ref{Fig.6}(c)-(k). LRMR, L1HyMixDe, SDeCNN, and 3DLogTNN are difficult to eliminate the deadlines in Fig.\ref{Fig.7}(c,g,h,j), and even cause a certain degree of degradation when dealing with strong noise. The denoised images obtained by LRTDTV and SNLRSF are oversmooth, some details in the original image are lost, and the edges of graphics become blurred in Fig.\ref{Fig.7}(e,i). NMoG introduces various levels of artifacts to the denoised image and blurring some details in Fig.\ref{Fig.7}(f). Therefore, the capability in removing mixed noise of the SWLRTR is proved from a visual perspective.

\subsubsection{ Running time }
Table III demonstrates the improvement of the proposed method in terms of computational complexity indicating the average computational time (second) of different methods for the three simulated datasets. The fastest results are highlighted in bold. The proposed SWLRTR has the shortest implementation time on the PaviaC dataset. On the other two datasets, SWLRTR is slower than LRMR, LRTV, and SDeCNN. The proposed method has low complexity while ensuring denoising performance.

\subsection{ Real data experiment }
In this section, we design real datasets experiments to prove the effectiveness of the SWLRTR on real-world noise. The Kennedy Space Center (KSC) dataset is size of ${\text{512}} \times {\text{614}} \times {\text{176}}$ after removing bands with poor quality. The Indian Pines dataset includes ${\text{145}} \times {\text{145}}$ pixels and 224 spectral bands. The two hyperspectral images are presented in Fig.\ref{Fig.8}. The aforementioned two datasets are employed in our real data experiments.

\subsubsection{ KSC dataset }
Due to the real dataset has no ground truth image as reference, we use the classification results provided by the radial basis function (RBF) kernel support vector machine to evaluate the performance of SWLRTR on the real dataset quantitatively. The KSC dataset contains 13 different types of land coverings that occur in the environment. Three evaluation indicators are listed in Table II namely the accuracy of each class, overall accuracy (OA), average accuracy (AA), and Kappa statistic. The best results are in bold. We can observe the proposed SWLRTR achieves the best results in all three measures compared with other methods which indicating the effectiveness in real image denoising. The denoising results on bands 89 and 50 are represented in Fig.\ref{Fig.9} and \ref{Fig.10}, respectively. From the figures, LRMR, SDeCNN, and SNLRSF can not restore the image well (Fig.\ref{Fig.9}), LRTV causes the image to be oversmooth, LRTDTV, NMoG L1HyMixDe, and 3DLogTNN tend to weaken the edges and details (Fig.\ref{Fig.10}).

\subsubsection{ Indian Pines dataset }
In the Indian Pines dataset, we show the restored HSIs of band 1 recovered by all competing methods in Fig.\ref{Fig.11} as an instance, because band 1 includes obvious impulse noise and other unknown noise. From Fig.\ref{Fig.11}(b,d), we can see LRMR and LRTDTV do not remove the impulse noise completely; LRTV, L1HyMixDe, and 3DLogTNN introduce the spectral distortion artifact; the result of NMoG is oversmooth; SDeCNN and SNLRSF have better performance than NMoG, however, still miss lots of details; as for the result of SWLRTR, the noise is removed satisfactorily and the different structural edges are preserved well. The validity of the proposed method is further proved by vertical mean profiles of band 28 in Fig.\ref{Fig.12}. The profiles are drawn according to the mean value of pixels in each column. Due to the presence of noise, the curve in the figure shows rapid fluctuations before denoising, and the fluctuations are suppressed by different methods after the restoration processing. From Fig.\ref{Fig.12} (j), the curve of SWLRTR method is smoother than others, which follows the visual results presented in Fig.\ref{Fig.11}.

\begin{figure}
\label{fig}
    \centering
    \subfigure[]{
        \includegraphics[width=0.46\linewidth]{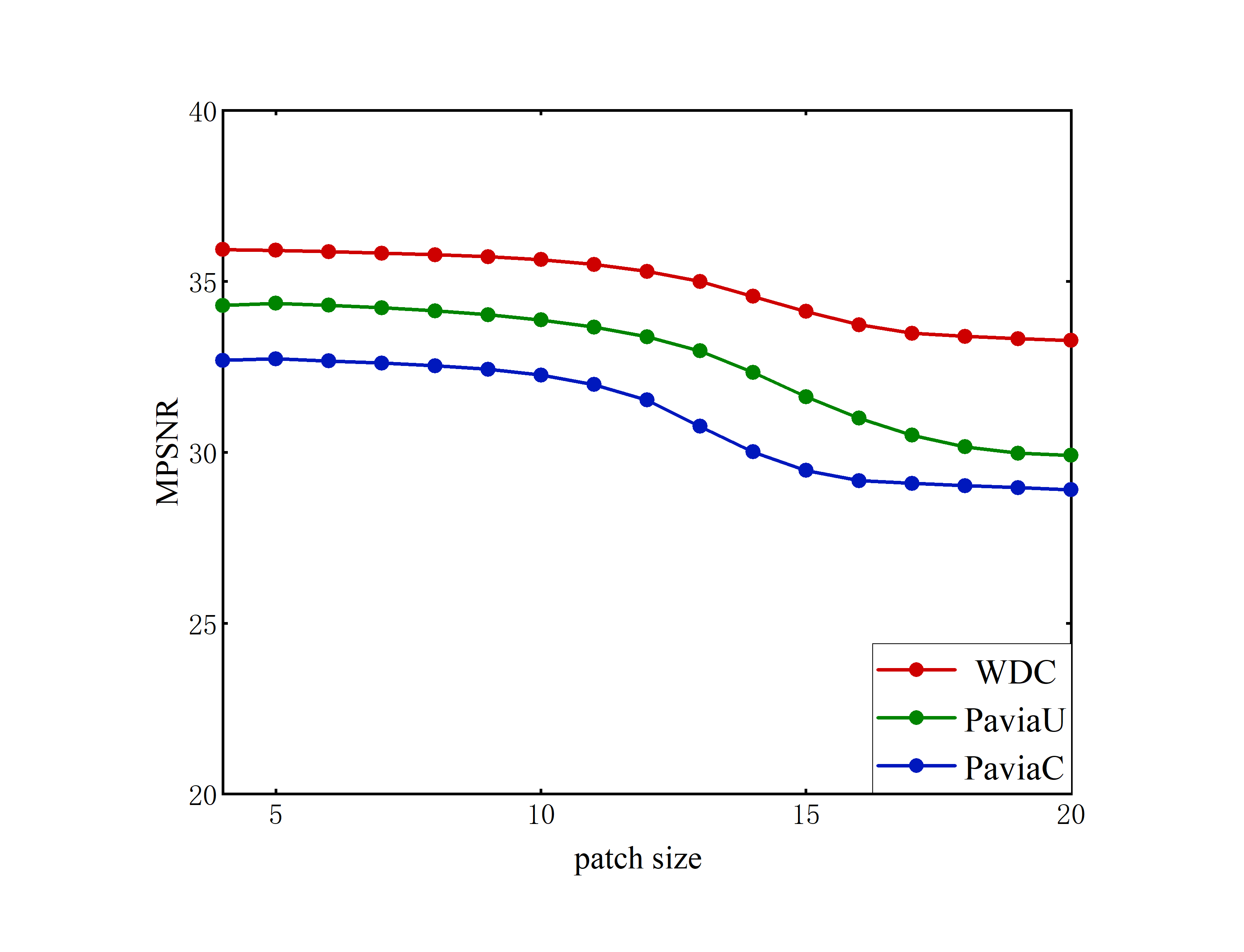}
    }
    \subfigure[]{
        \includegraphics[width=0.46\linewidth]{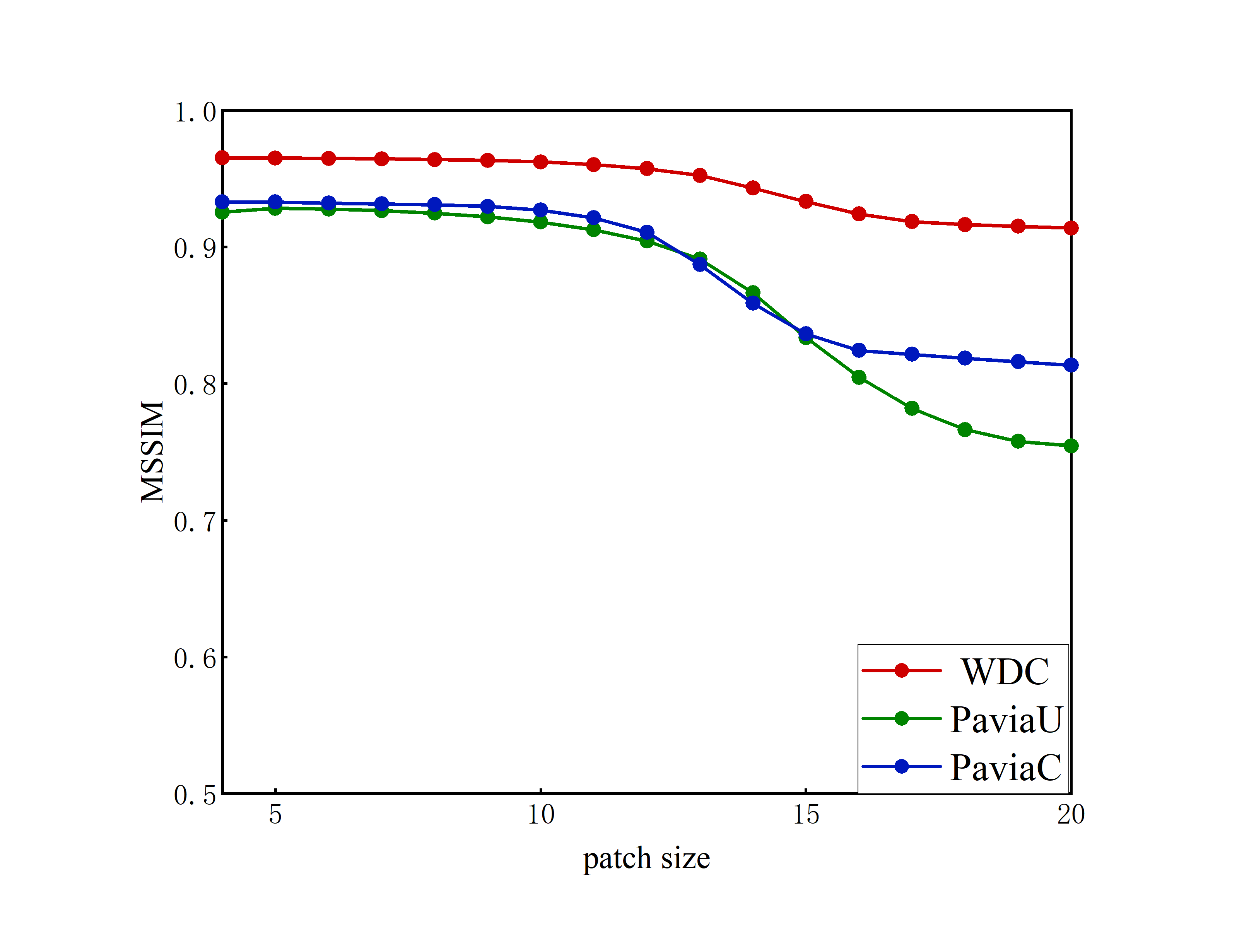}
    }
    \centering
    \caption{ Sensitivity analysis of patch size }
    \label{Fig.13}
\end{figure}

\begin{figure}
    \centering
    \subfigure[]{
        \includegraphics[width=0.46\linewidth]{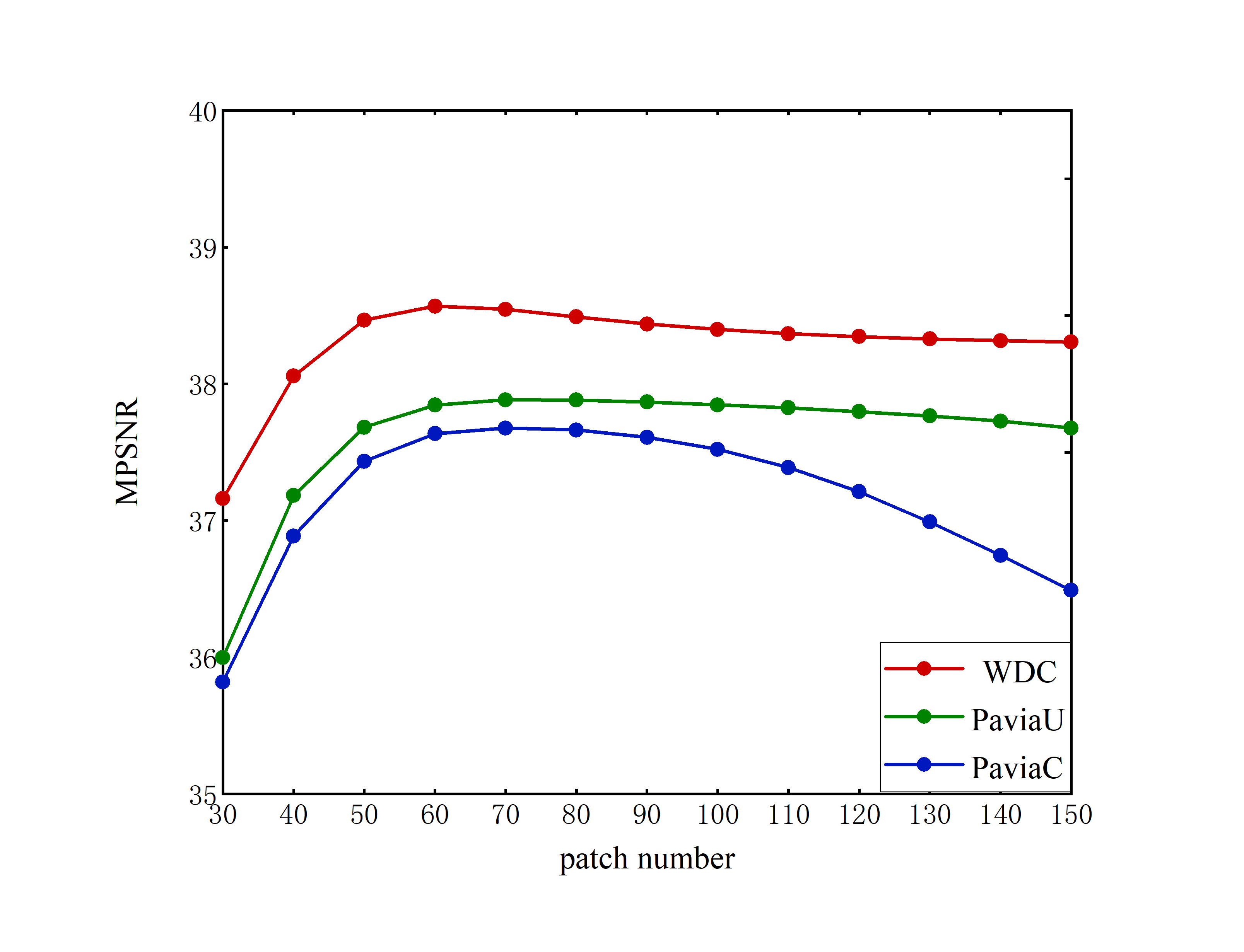}
    }
    \subfigure[]{
        \includegraphics[width=0.46\linewidth]{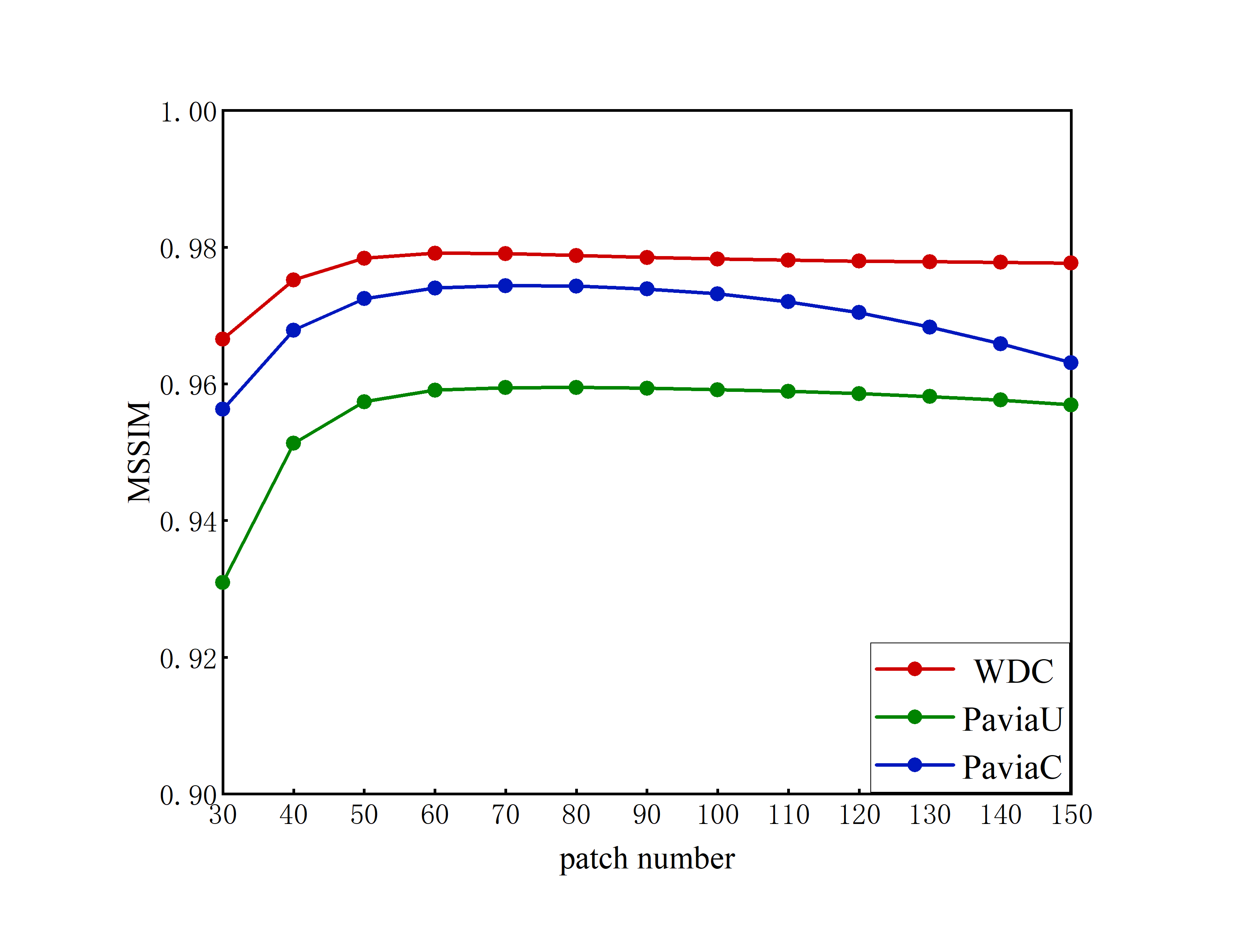}
    }
    \centering
    \caption{ Sensitivity analysis of patch number }
    \label{Fig.14}
\end{figure}

\begin{figure}
    \centering
    \subfigure[]{
        \includegraphics[width=0.46\linewidth]{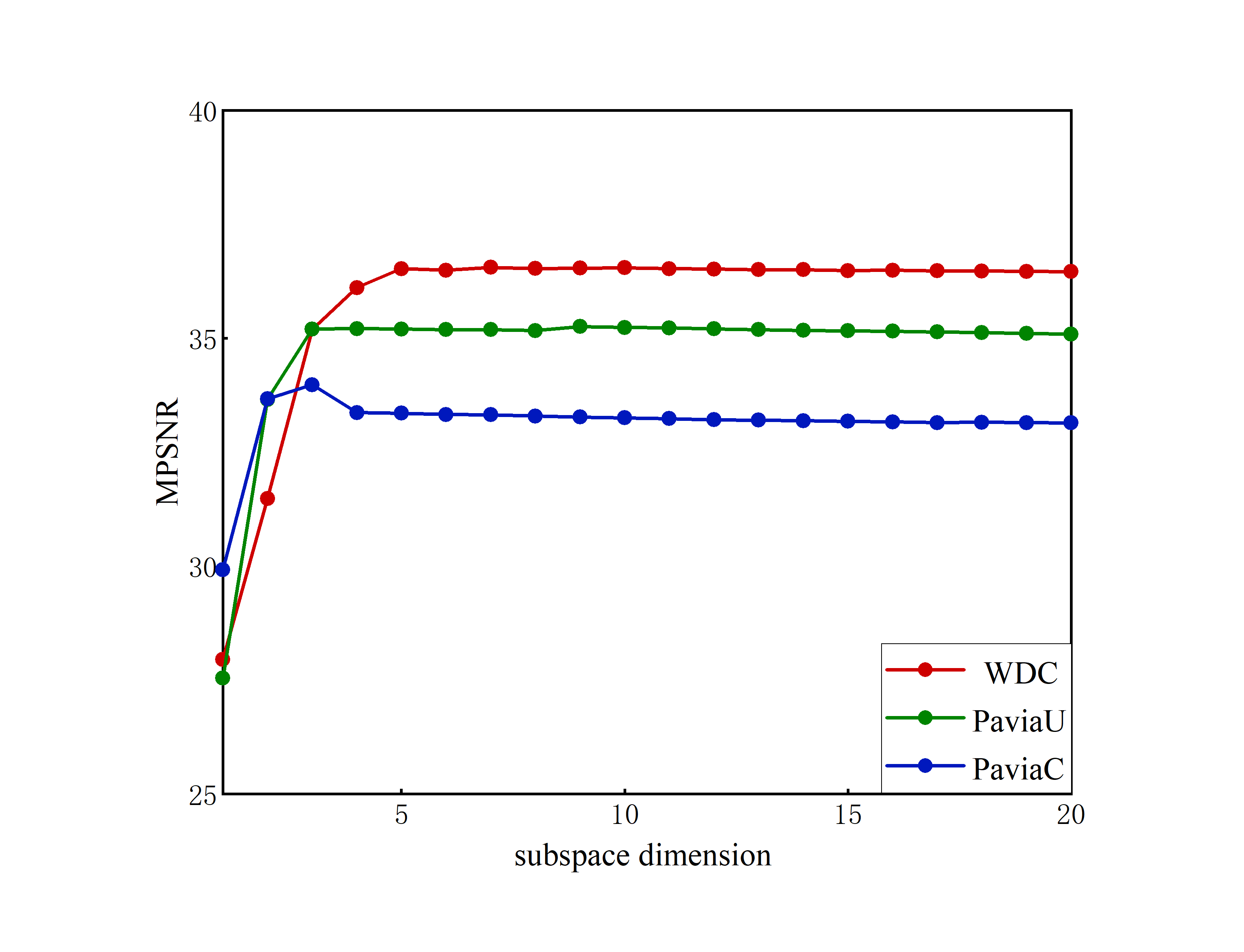}
    }
    \subfigure[]{
        \includegraphics[width=0.46\linewidth]{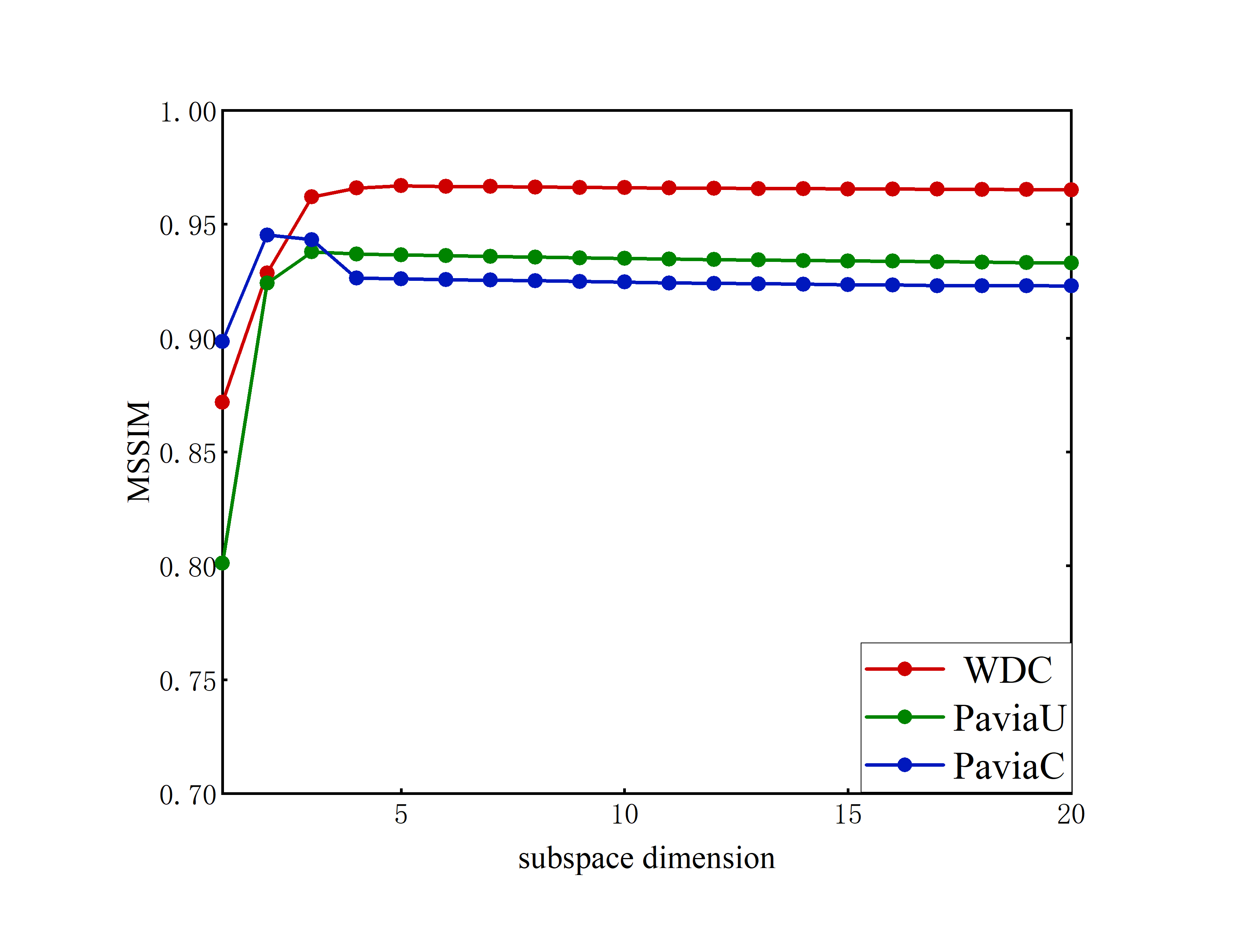}
    }
    \centering
    \caption{ Sensitivity analysis of subspace dimension }
    \label{Fig.15}
\end{figure}

\begin{figure}
    \centering
    \subfigure[]{
        \includegraphics[width=0.46\linewidth]{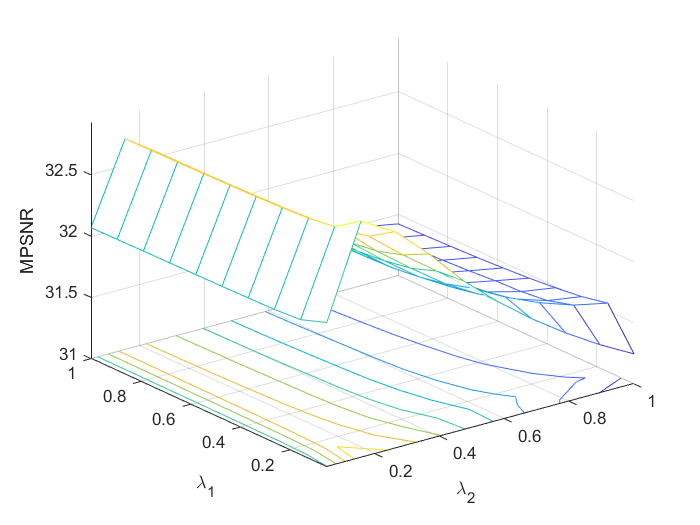}
    }
    \subfigure[]{
        \includegraphics[width=0.46\linewidth]{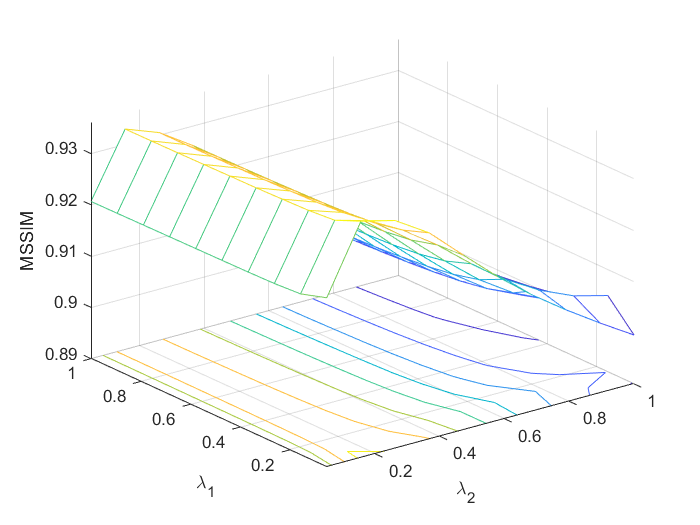}
    }
    \centering
    \caption{ Sensitivity analysis of regularization parameters ${\lambda _{1}}$ and ${\lambda _{2}}$ }
    \label{Fig.16}
\end{figure}

\subsection{ Sensitivity analysis }
The proposed SWLRTR contains five important parameters, including patch size p, the number of similar patches q, subspace dimension k, and regularized parameters ${\lambda _{1}}$, ${\lambda _{2}}$. In this section, we experimentally explore the optimal settings of these parameters. Besides, the experiments are implemented in the simulated datasets of case 4.

\subsubsection{ patch size p}
By changing p from 3 to 20, Fig.\ref{Fig.13} shows the denoising results of MPSNR and MSSIM. The step size is set to 3. As we can see, the denoising results are relatively stable for $3 < p < 10$, then appears degradation when $p > 10$. When the selected patch is too small, the local spatial information within the image will be lost and not conducive to block matching, when the selected patch is too large, the detail information will be eliminated. In addition, with the increase of patch size, the amount of calculation will also increase. In order to have better denoising ability and lower computational complexity, the patch size is set to 5 in our experiments.

\subsubsection{ the number of similar patches q }
The number of similar patches selected in the block matching stage is also an important parameter that affects the results of image denoising. In the sensitivity experiment, the parameter q is selected from 30 to 150. Fig.\ref{Fig.14} presents the corresponding MPSNR and MSSIM values where the highest MPSNR and MSSIM are achieved at $50 < q < 70$ for different datasets. The main reason is that either too large or too small number of similar patches can lead to insufficient use of nonlocal self-similarity. Hence, we choose the similar patch number q according to the noise intensity, fewer patches are selected when the noise is heavy, vice versa. In this paper, q is set to 70 when the noise variance less than or equal to 0.1, otherwise, q is set to 70.

\subsubsection{ subspace dimension k }
Generally, the optimal subspace dimension is distinct for different datasets. The proposed method achieves the best denoising result when the subspace dimension is 5, 4, 3 for WDC, PaviaU, and PaviaC dataset, respectively, as presented in Fig.\ref{Fig.15}. In the real data experiment, the subspace dimension is set to 10, 25 for KSC and Indian Pines dataset. Besides, the subspace dimension is manually adjusted in our experiments, which is usually different from the value obtained by the HySime algorithm, because the algorithm only considers Gaussian noise leading to an inaccurate estimate.

\subsubsection{ regularized parameter ${\lambda _{1}}$, ${\lambda _{2}}$ }
By varying ${\lambda _{1}}$ and ${\lambda _{2}}$ from 0 to 1 with a step size of 0.05, Fig.\ref{Fig.16} shows the denoising results under MPSNR and MSSIM measurements. Therefore, in order to achieve robust performance, the parameter ${\lambda _{1}}$ and ${\lambda _{2}}$ are set to 0.2, 0.1, respectively.

\section{ Conclusion }
In this paper, a novel hyperspectral image denoising model integrated the weighted low-rank tensor regularization term and subspace representation for mixed noise is proposed. Based on the fact that HSI exists in multiple subspaces, each spectral signature can be represented by a linear combination of a few pure spectral endmembers. The global spectral low rankness can be enforced by decomposing the HSI into mode-3 tensor-matrix product of a reduced image and a low-dimensional orthogonal matrix. The experimental results show that the computational cost is greatly reduced because of the subspace representation. To further utilize the priors in HSI, a low-rank tensor is constructed by aggregating the similar patches searched from the reduced image, then we introduce the weighted low-rank tensor regularizer to exploit its low rankness. Specifically, the regularization punishes the elements in the core tensor reasonably. Besides, we design an algorithm for solving the model based on alternate minimization. Experiments on simulated and real datasets show that the proposed algorithm has better denoising performance compared with other advanced algorithms.


%



\ifCLASSOPTIONcaptionsoff
  \newpage
\fi



%

\bibliographystyle{IEEEtran}
\bibliography{SWLRTR}

%




\end{document}